\documentclass[10pt,twocolumn,letterpaper]{article}

\usepackage{cvpr}
\usepackage{times}
\usepackage{epsfig}
\usepackage{graphicx}
\usepackage{amsmath}
\usepackage{amssymb}
\usepackage{adjustbox}
\usepackage{array}
\usepackage{subcaption}
\usepackage{color}
\usepackage{cite}
\usepackage{float}
\usepackage{url}

\newcommand{\figref}[1]{Figure~\ref{#1}}

\newcommand{\eqnref}[1]{Equation~(\ref{#1})}

\newcommand{\secref}[1]{Section~\ref{#1}}

\newcolumntype{M}[1]{>{\centering\arraybackslash}m{#1\linewidth}}
\newcolumntype{C}[1]{>{\centering\arraybackslash}p{#1\linewidth}}

\usepackage[pagebackref=true,breaklinks=true,letterpaper=true,colorlinks,bookmarks=false]{hyperref}

\cvprfinalcopy 

\def\cvprPaperID{645} 

\ifcvprfinal\fi
\begin{document}

\title{A Unified Approach of Multi-scale Deep and Hand-crafted Features for Defocus Estimation}

\author{
Jinsun Park$^{\dagger}$\\
{\tt\small zzangjinsun@gmail.com}
\and
Yu-Wing Tai$^{\ddagger}$\\
{\tt\small yuwing@gmail.com}
\and
Donghyeon Cho$^{\dagger}$\\
{\tt\small cdh12242@gmail.com}
\and
In So Kweon$^{\dagger}$\\
{\tt\small iskweon@kaist.ac.kr}
\and
$^{\dagger}$Robotics and Computer Vision Lab., KAIST, Korea, Republic of\\
$^{\ddagger}$Tencent YouTu Lab., China
}



\maketitle

\begin{abstract}
In this paper, we introduce robust and synergetic hand-crafted features and a simple but efficient deep feature from a convolutional neural network (CNN) architecture for defocus estimation. This paper systematically analyzes the effectiveness of different features, and shows how each feature can compensate for the weaknesses of other features when they are concatenated. For a full defocus map estimation, we extract image patches on strong edges sparsely, after which we use them for deep and hand-crafted feature extraction. In order to reduce the degree of patch-scale dependency, we also propose a multi-scale patch extraction strategy. A sparse defocus map is generated using a neural network classifier followed by a probability-joint bilateral filter. The final defocus map is obtained from the sparse defocus map with guidance from an edge-preserving filtered input image. Experimental results show that our algorithm is superior to state-of-the-art algorithms in terms of defocus estimation. Our work can be used for applications such as segmentation, blur magnification, all-in-focus image generation, and 3-D estimation.
\end{abstract}
\vspace{-0.15in}

\section{Introduction}
\label{sec:Introduction}

The amount of defocus represents priceless information can be obtained from a single image. If we know the amount of defocus at each pixel in an image, higher level information can be inferred based on defocus values such as depth~\cite{ziou2001depth}, salient region~\cite{jiang2013salient} and foreground and background of a scene~\cite{mcguire2005defocus} and so on. Defocus estimation, however, is a highly challenging task, not only because the estimated defocus values vary spatially, but also because the estimated solution contains ambiguities~\cite{Levin2007tog}, where the appearances of two regions with different amounts of defocus can be very similar. Conventional methods~\cite{bae2007defocus, tai2009single, zhuo2011defocus} rely on strong edges to estimate the amount of defocus. Determining the amount of defocus only based on the strength of strong edges, however, may lead to overconfidence and misestimations. Thus, we need a more reliable and robust defocus descriptor for defocus estimations.

In this paper, we present hand-crafted and deep features which assess various aspects of an image for defocus estimation, and a method to obtain a reliable full defocus map of a scene. Our hand-crafted features focus on three components of an image: the frequency domain power distribution, the gradient distribution and the singular values of an image. We also utilize a convolutional neural network (CNN) to extract high-dimensional deep features directly learnt from millions of in-focus and blurred image patches. All of the features are concatenated to construct our defocus feature vector and are fed into a fully connected neural network classifier to determine the amount of defocus.

\begin{figure}
\begin{center}
\begin{tabular}{@{}c@{\hskip 0.005\linewidth}c@{\hskip 0.005\linewidth}c}
\includegraphics[width=0.330\linewidth]{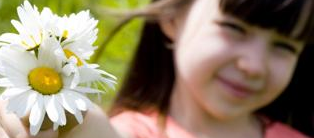} &
\includegraphics[width=0.330\linewidth]{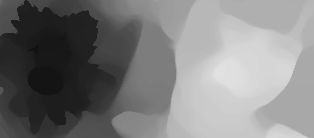} &
\includegraphics[width=0.330\linewidth]{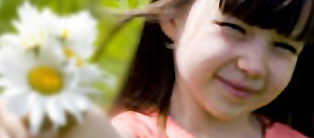} \\
{\small (a) Image} & {\small (b) Defocus Map} & {\small (c) Refocused}
\end{tabular}
\end{center}
\vspace{-0.25in}
\caption{Our defocus estimation result and a digital refocusing application example.}
\vspace{-0.20in}
\label{fig:Teaser}
\end{figure}

One of the challenges associated with the defocus estimation is the vagueness of the amount of defocus in homogeneous regions, as such regions show almost no difference in appearance when they are in-focus or blurred. To avoid this problem, we first estimate the amount of defocus using multi-scale image patches from only strong edges, and then propagate the estimated values into homogeneous regions with the guidance of edge-preserving filtered input image. The full defocus map is obtained after the propagation step. We use the defocus map for various applications, such as segmentation, blur magnification, all-in-focus image generation, and 3-D estimation. \figref{fig:Teaser} shows an example of our defocus estimation result and a refocusing application.

\section{Related Work}
\label{sec:RelatedWork}

\textbf{Defocus Estimation} \ Defocus estimation plays an important role in many applications in the computer vision community. It is used in digital refocusing~\cite{bae2007defocus, cao2013digital}, depth from defocus~\cite{lin2013absolute, ziou2001depth, shi2015break}, salient region detection~\cite{jiang2013salient} and image matting~\cite{mcguire2005defocus}, to name just a few. 

Elder and Zucker~\cite{elder1998local} estimate the minimum reliable scale for edge detection and defocus estimation. They utilize second derivative Gaussian filter responses, but this method is not robust due to errors which arise during the localization of edges. Bae and Durand~\cite{bae2007defocus} also utilize second derivative Gaussian filter responses to magnify the amount of defocus on the background region. However, their strategy is time-consuming owing to its use of a brute-force scheme. Tai and Brown~\cite{tai2009single} employ a measure called \textit{local contrast prior}, which considers the relationships between local image gradients and local image contrasts, but the local contrast prior is not robust to noise. Zhuo and Sim~\cite{zhuo2011defocus} use the ratio between the gradients of input and re-blurred images with a known Gaussian blur kernel. However, it easily fails with noise and edge mislocalization. Liu \emph{et al.}~\cite{liu2008image} inspect the power spectrum slope, gradient histogram, maximum saturation and autocorrelation congruency. Their segmentation result, however, cannot precisely localize blurry regions. Shi \emph{et al.}~\cite{shi2014discriminative} use not only statistical measures such as peakedness and heavy-tailedness but also learnt filters from image data with labels. Homogeneous regions in the image are weak points in their algorithm. Shi \emph{et al.}~\cite{shi2015just} construct sparse dictionaries containing sharp and blurry bases and determine which dictionary can reconstruct an input image sparsely, but their algorithm is not robust to large blur, as it is tailored for just noticeable blur estimation.

\textbf{Neural Networks} \ Neural networks have proved their worth as algorithms superior to their conventional counterparts in many computer vision tasks, such as object and video classification~\cite{karpathy2014large, ILSVRC15}, image restoration~\cite{Eigen2013iccv}, image matting~\cite{cho2016natural}, image deconvolution~\cite{xu2014deep}, motion blur estimation~\cite{sun2015learning}, blur classification~\cite{aizenberg2008blur, yan2013image}, super-resolution~\cite{dong2014learning}, salient region detection~\cite{lee2016saliency} and edge-aware filtering~\cite{xu2015deep}.

Sun \emph{et al.}~\cite{sun2015learning} focus on motion blur kernel estimation. They use a CNN to estimate pre-defined discretized motion blur kernels. However, their approach requires rotational input augmentation and takes a considerable amount of time during the MRF propagation step. Aizenberg \emph{et al.}~\cite{aizenberg2008blur} use a multilayer neural network based on multivalued neurons (MVN) for blur identification. The MVN learning step is computationally efficient. However, their neural network structure is quite simple. Yan and Shao~\cite{yan2013image} adopt two-stage deep belief networks (DBN) to classify blur types and to identify blur parameters, but they only rely on features from the frequency domain.

\textbf{Our Work} \ Compared with the previous works, instead of using only hand-crafted features, we demonstrate how we can apply deep features to the defocus estimation problem. The deep feature is learnt directly from training data with different amounts of defocus blur. Because each extracted deep feature is still a local feature, our hand-crafted features, which capture both local and global information of an image patch, demonstrate the synergetic effect of boosting the performance of our algorithm. Our work significantly outperforms previous works on defocus estimation in terms of both quality and accuracy.

\section{Feature Extraction}
\label{sec:FeatureExtraction}

We extract multi-scale image patches from an input image for feature extraction. In addition, we extract image patches on edges only because homogeneous regions are ambiguous in defocus estimation. For edge extraction, we first transform the input image from the RGB to the HSV color space and then use a V channel to extract image edges.

There have been numerous hand-crafted features for sharpness measurements~\cite{batten2000autofocusing, marichal1999blur, wee2008image, tai2009single, liu2008image}. In this work, three hand-crafted features related to the frequency domain power distribution, the gradient distribution, and the singular values of a grayscale image patch are proposed. The deep feature is extracted from a CNN which directly processes color image patches in the RGB space for feature extraction. All of the extracted features are then concatenated to form our final defocus feature.

\begin{figure}
\begin{center}
\begin{tabular}{@{}c@{\hskip 0.01\linewidth}c@{\hskip 0.01\linewidth}c}
\includegraphics[width=0.270\linewidth]{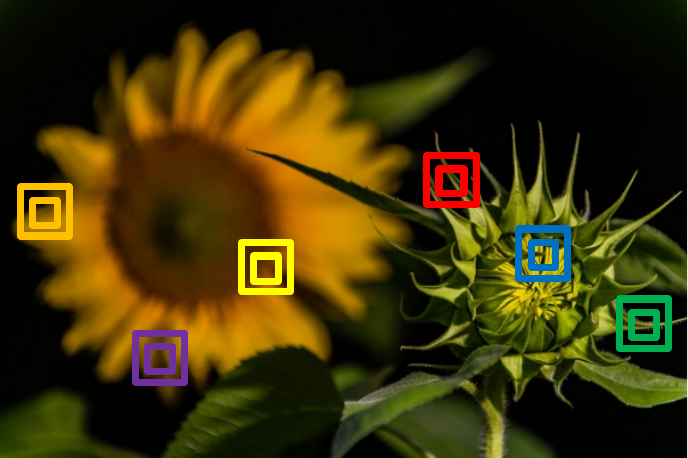} &
\includegraphics[width=0.330\linewidth]{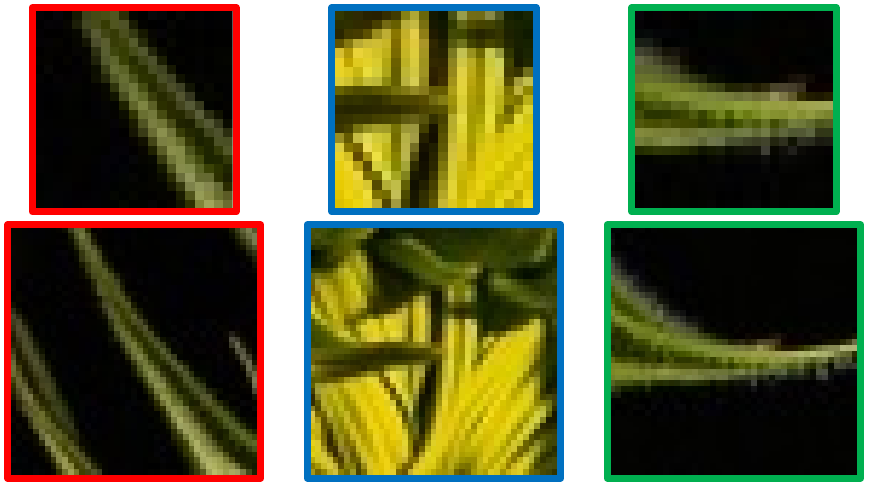} &
\includegraphics[width=0.330\linewidth]{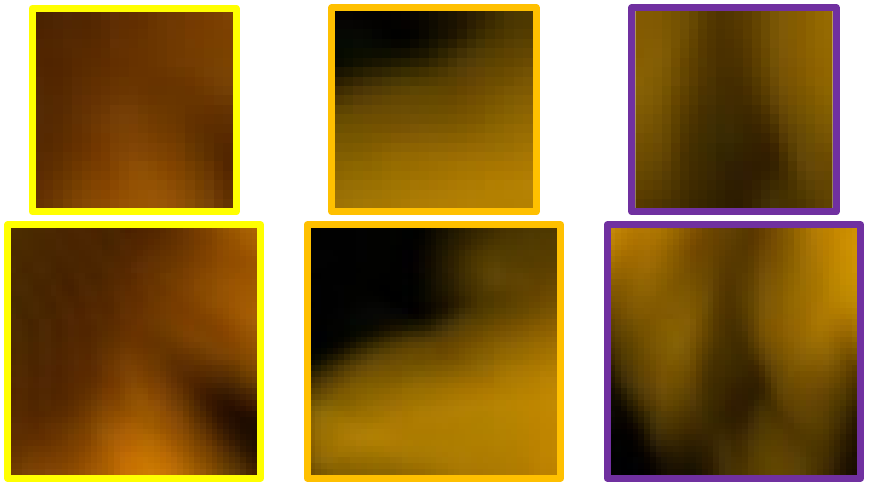} \\
{\small (a) Image} & {\small (b) From strong edges} & {\small (c) From weak edges}
\end{tabular}
\end{center}
\vspace{-0.2in}
\caption{Multi-scale patches from strong and weak edges.}
\label{fig:MultiScalePatches}
\vspace{-0.2in}
\end{figure}

\begin{figure*}
\begin{center}
\begin{tabular}{@{}c@{\hskip 0.01\linewidth}c@{\hskip 0.01\linewidth}c}
\includegraphics[width=0.320\linewidth]{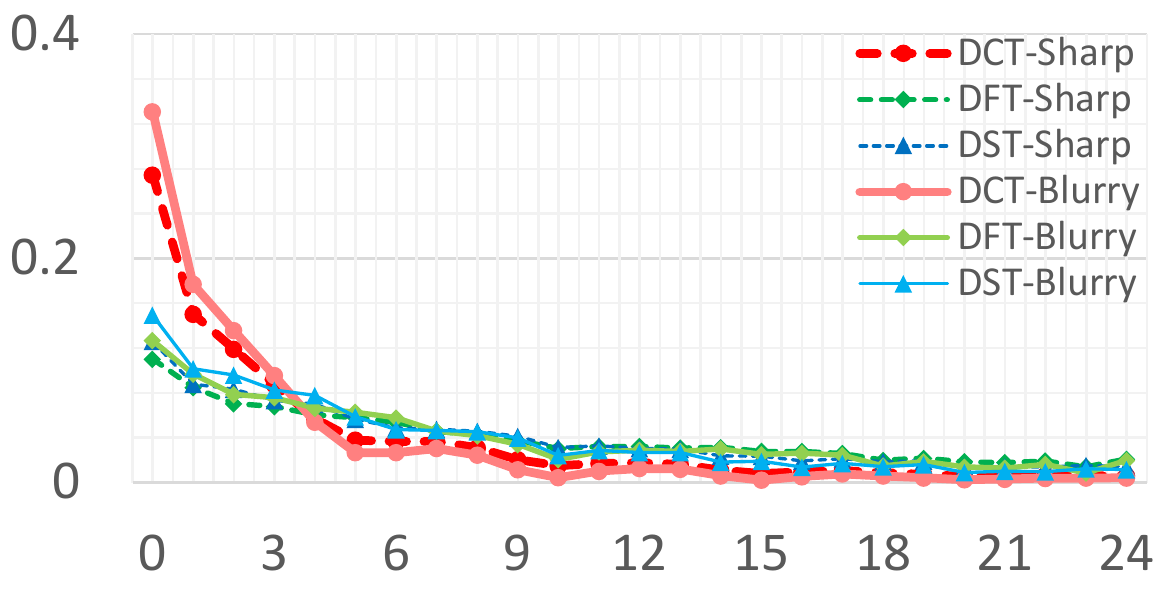} &
\includegraphics[width=0.320\linewidth]{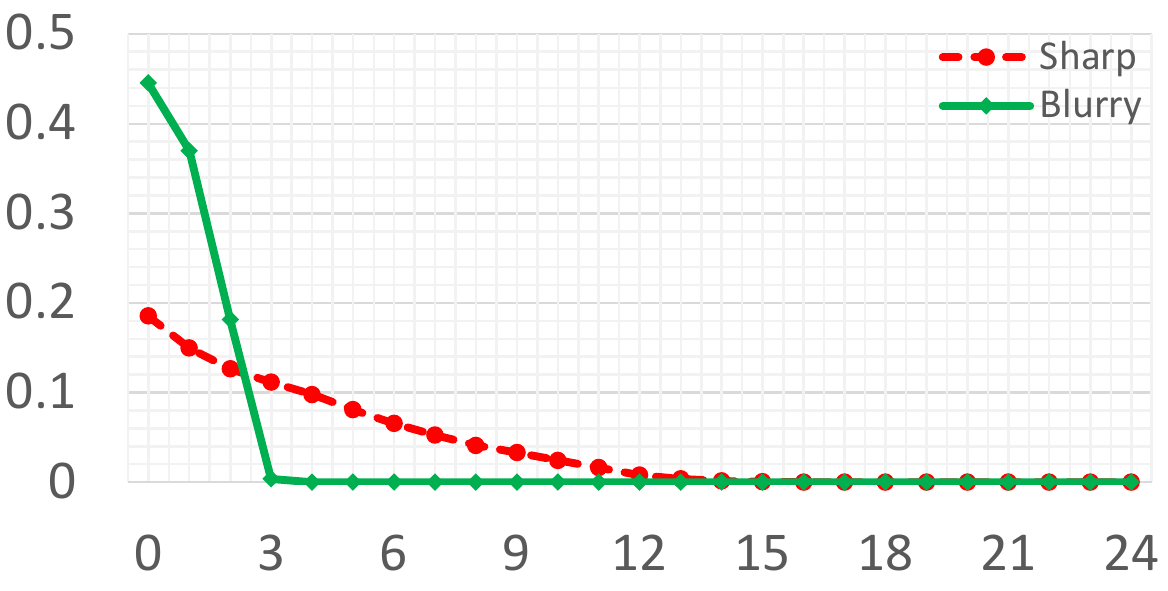} &
\includegraphics[width=0.320\linewidth]{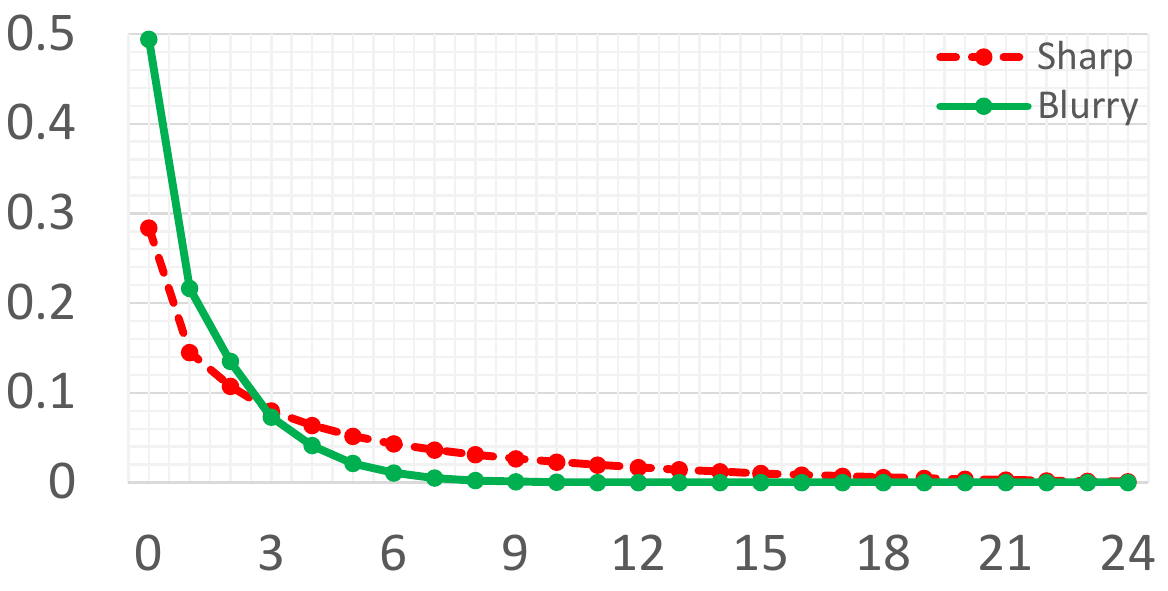} \\
{\small (a)} & {\small (b)} & {\small (c)}
\end{tabular}
\end{center}
\vspace{-0.25in}
\caption{Sharp and blurry hand-crafted features. Average (a) DCT, DFT, DST, (b) gradient and (c) SVD features from sharp (dotted) and blurry (solid) patches. The number of samples exceeds more than 11K. The absolute difference between sharp and blurry features can be a measure of discriminative power.}
\label{fig:HandCraftedFeatures}
\vspace{-0.2in}
\end{figure*}

\subsection{Multi-scale Patch Extraction}
Because we extract hand-crafted and deep features based on image patches, it is important to determine a suitable patch size for each pixel in an image. Although there have been many works related to scale-space theories~\cite{lindeberg1998feature, liu2012scale, lowe2004distinctive}, there is still some ambiguity with regard to the relationship between the patch scale and the hand-crafted features, as they utilize global information in an image patch. In other words, a sharp image patch can be regarded as blurry mistakenly depending on the size of the patch, and vice versa. In order to avoid patch scale dependency, we extract multi-scale patches depending on the strength of the edges. In natural images, strong edges are more likely to be in-focus than blurry ones ordinarily. Therefore, we assume that image patches from strong edges are in-focus and that weak edges are blurry during the patch extraction step. For sharp patches, we can determine their sharpness accurately with a small patch size, whereas blurry patches can be ambiguous with a small patch size because of their little change in the appearance when they are blurred or not. \figref{fig:MultiScalePatches} shows multi-scale patches from strong and weak edges. \figref{fig:MultiScalePatches} (b) shows that small patches from strong edges still have abundant information for defocus estimation, while \figref{fig:MultiScalePatches} (c) shows that small patches from weak edges severely lack defocus information and contain high degrees of ambiguity. Based on this observation, we extract small patches from strong edges and large patches from weak edges. Edges are simply extracted using the Canny edge detector~\cite{canny1986computational} with multi-threshold values. In general, the majority of the strong edges can be extracted in an in-focus area. Weak edges can be extracted in both sharp and blurry areas because they can come from in-focus weak textures or out-of-focus sharp textures. Our multi-scale patch extraction scheme boosts the performance of a defocus estimation algorithm drastically (\secref{subsec:defocusfeature}).

\subsection{DCT Feature}
We transform a grayscale image patch $P_{I}$ to the frequency domain to analyze its power distribution. We utilize the discrete cosine transform (DCT) because the DCT offers strong energy compaction~\cite{rao2014discrete}; i.e., most of the information pertaining to a typical signal tends to be concentrated in a few low-frequency bands. Hence, when an image is more detailed, more non-zero DCT coefficients are needed to preserve the information. Accordingly, we can examine high-frequency bands at a higher resolution with the DCT than with the discrete Fourier transform (DFT) or the discrete sine transform (DST). Because an in-focus image has more high-frequency components than an out-of-focus image, the ratio of high-frequency components in an image patch can be a good measure of the blurriness of an image. Our DCT feature $f_{D}$ is constructed using the power distribution ratio of frequency components as follows:
\begin{equation}
f_{D}(k) = \frac{1}{W_{D}}\log\left({1+\sum_{\theta}\sum_{\rho = \rho_{k}}^{\rho_{k+1}}\frac{|\mathcal{P}(\rho,\theta)|}{S_{k}}}\right), \ k \in [1, n_{D}],
\label{eq:fDCT}
\end{equation}
where $|\cdot|$, $\mathcal{P}(\rho,\theta)$, $\rho_{k}$, $S_{k}$, $W_{D}$ and $n_{D}$ denote the absolute operator, the discrete cosine transformed image patch with polar coordinates, the $k$-th boundary of the radial coordinate, the area enclosed by $\rho_{k}$ and $\rho_{k+1}$, a normalization factor to make sum of the feature unity, and the dimensions of the feature, respectively. \figref{fig:HandCraftedFeatures} (a) shows features from sharp and blurry patches after different transformations. The absolute difference between sharp and blurry features can be a measure of discriminative power. In this case, the DCT feature has the best discriminative power because its absolute difference between sharp and blurry features is greater than those of the other transformations.

\begin{figure}
\begin{center}
\begin{tabular}{@{}c@{\hskip 0.005\linewidth}c@{\hskip 0.005\linewidth}c@{\hskip 0.005\linewidth}c@{\hskip 0.005\linewidth}c}
\includegraphics[width=0.245\linewidth]{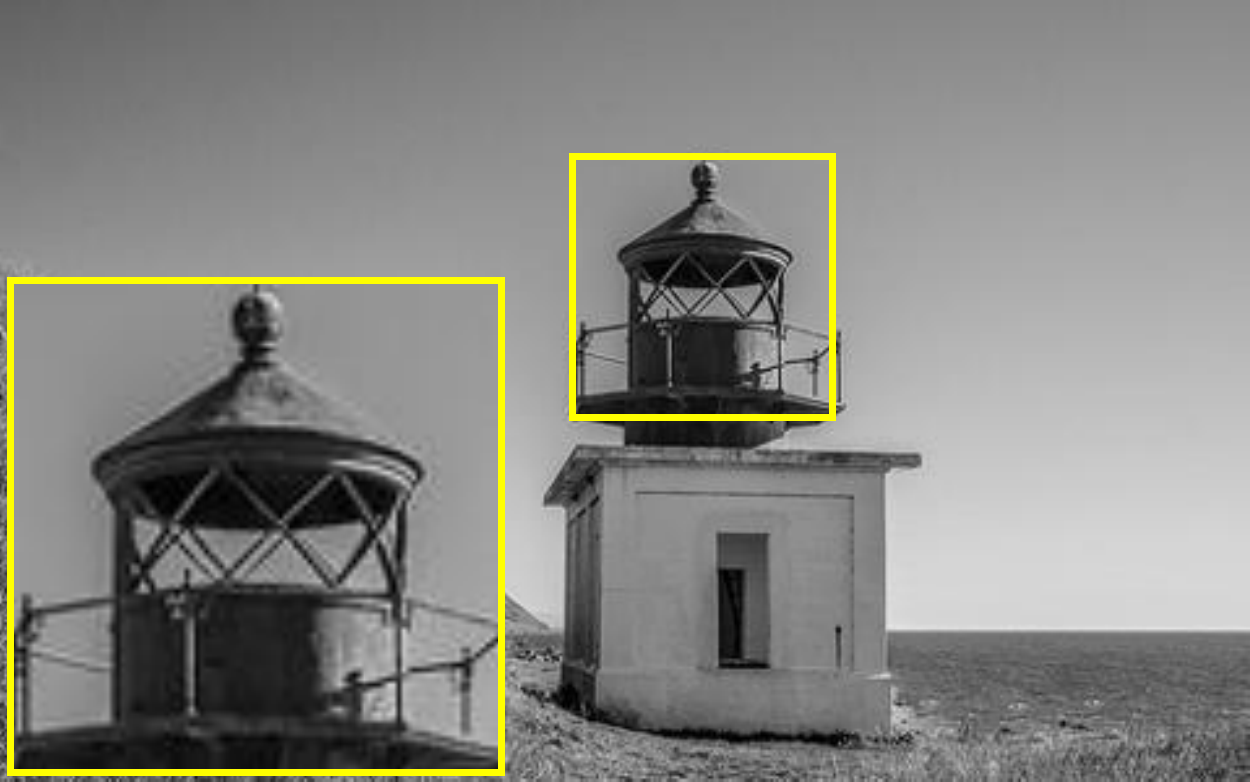} &
\includegraphics[width=0.245\linewidth]{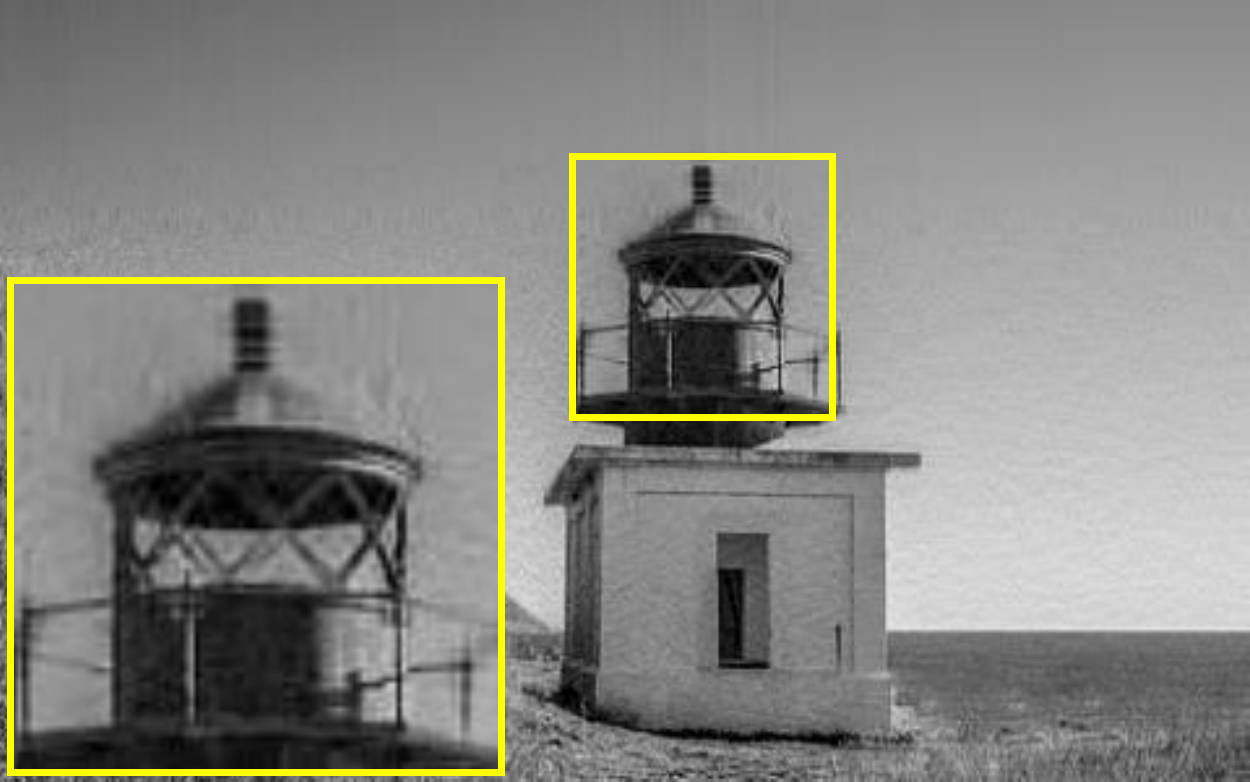} &
\includegraphics[width=0.245\linewidth]{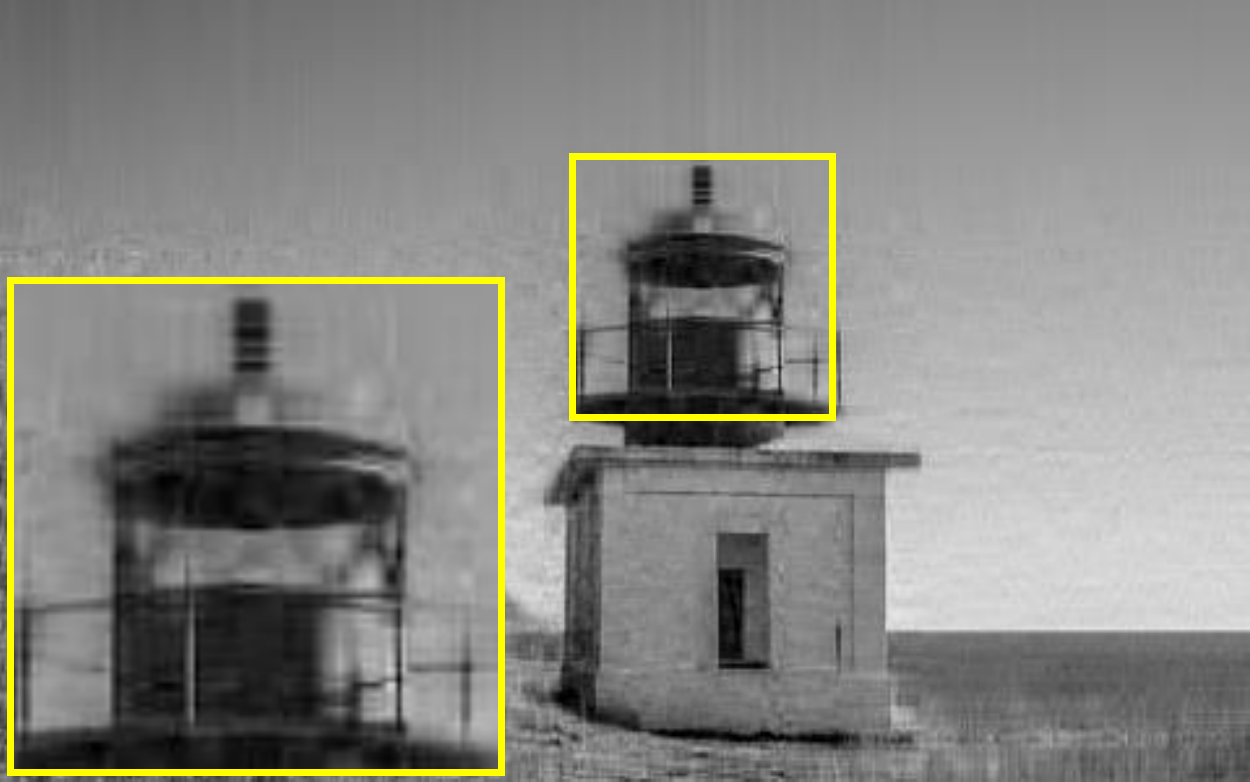} &
\includegraphics[width=0.245\linewidth]{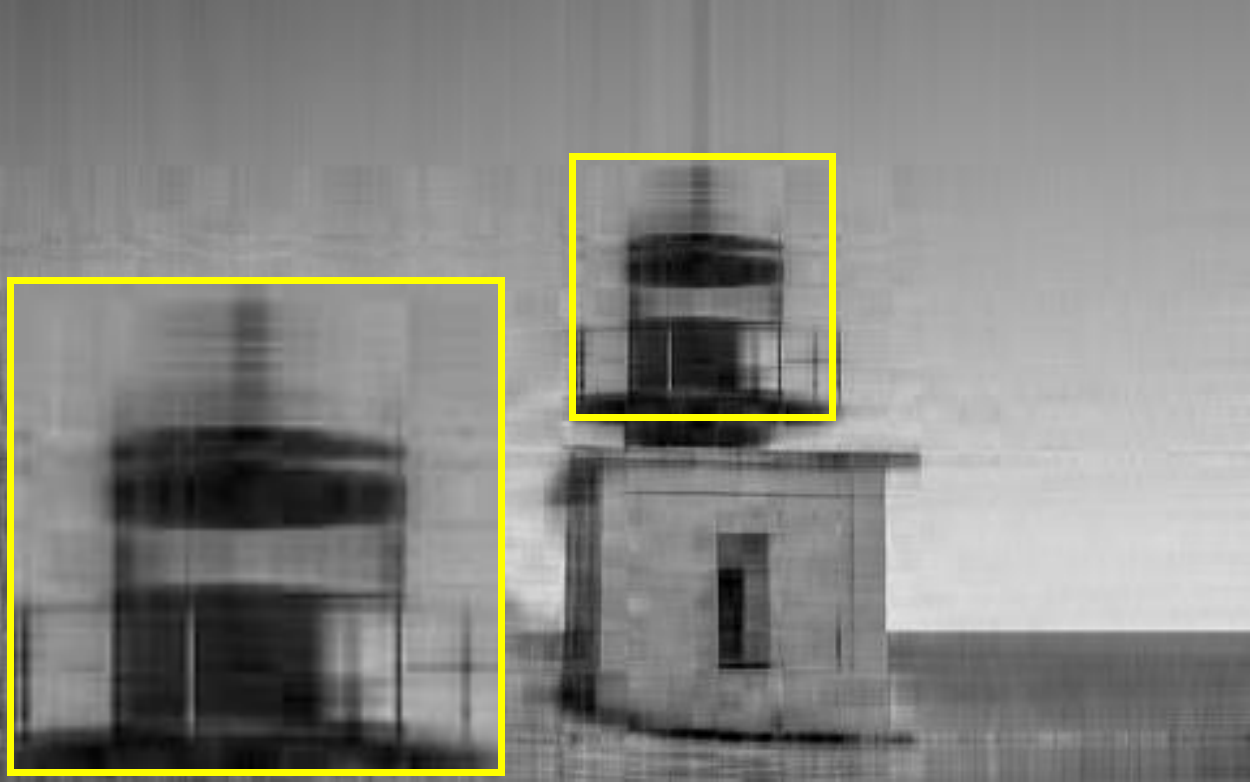} \\
{\small (a) Original} & {\small (b) $n = 60$} & {\small (c) $n = 30$} & {\small (d) $n = 15$}
\end{tabular}
\end{center}
\vspace{-0.25in}
\caption{Low-rank matrix approximation of an image. (a) Original and (b)-(d) approximated images with the number of preserved singular values. The more number of singular values are preserved, the more details are also preserved.}
\label{fig:LowrankMatrixApproximation}
\vspace{-0.2in}
\end{figure}

\begin{figure*}
\begin{center}
\begin{tabular}{@{}c}
\includegraphics[width=0.980\linewidth]{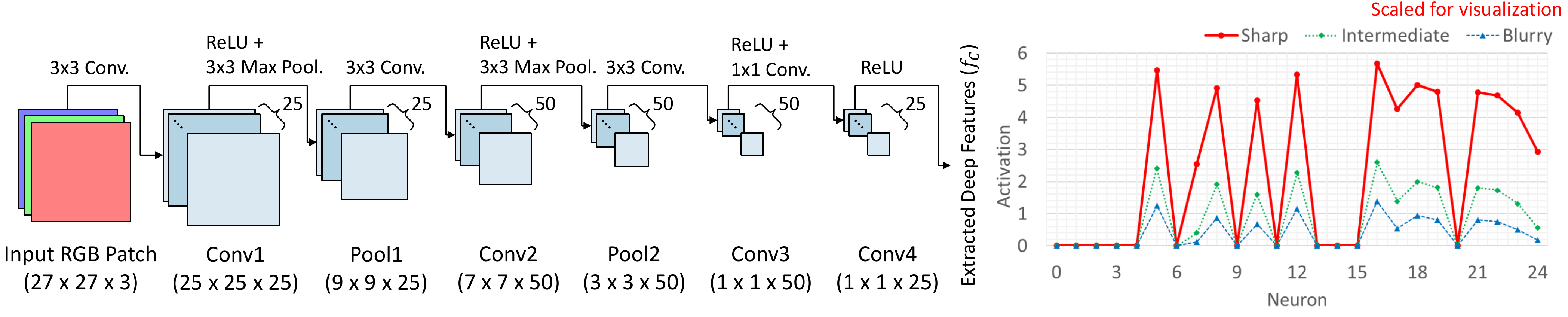}
\end{tabular}
\end{center}
\vspace{-0.3in}
\caption{Our deep feature extraction network and average activations with sharp, intermediate and blurry patches. The output dimensions of each layer are shown together. The stride is set to 1 for convolution and to 3 for max pooling.}
\label{fig:CNN}
\vspace{-0.2in}
\end{figure*}

\subsection{Gradient Feature}
We calculate the gradients of $P_{I}$ using Sobel filtering to obtain a gradient patch $P_{G}$. Typically, there are more strong gradients in a sharp image patch than in a blurry image. Therefore, the ratio of the strong gradient components in an image patch can be another measure of the sharpness of the image. We use the normalized histogram of $P_{G}$ as a second component of our defocus feature. Our gradient feature $f_{G}$ is defined as follows:
\begin{equation}
f_{G}(k) = \frac{1}{W_{G}}\log\left(1+\mathcal{H}_{G}(k)\right), \ k \in [1, n_{G}],
\label{eq:fGRD}
\end{equation}
where $\mathcal{H}_{G}$, $W_{G}$ and $n_{G}$ denote the histogram of $P_{G}$, the normalization factor and the dimensions of the feature, respectively. \figref{fig:HandCraftedFeatures} (b) shows a comparison of sharp and blurry gradient features. Despite its simplicity, the gradient feature shows quite effective discriminative power.

\subsection{SVD Feature}
Singular value decomposition (SVD) has many useful applications in signal processing. One such application of SVD is the low-rank matrix approximation~\cite{press2007numerical} of an image. The factorization of an $m \times n$ real matrix $\mathbf{A}$ can be written as follows:
\begin{equation}
\mathbf{A} = \mathbf{U \Lambda V}^{T} = \sum_{k=1}^{N}A_{k} = \sum_{k=1}^{N}\lambda_{k}u_{k}v_{k}^{T},
\label{eq:SVD}
\end{equation}
where $\mathbf{\Lambda}$, $N$, $\lambda_{k}$, $u_{k}$ and $v_{k}$ denote the $m \times n$ diagonal matrix, the number of non-zero singular values of $\mathbf{A}$, the $k$-th singular value, and the $k$-th column of the real unitary matrices $\mathbf{U}$ and $\mathbf{V}$, respectively. If we construct a matrix $\tilde{\mathbf{A}} = \sum_{k=1}^{n}A_{k}$, where $n \in [1, N]$, we can approximate the given matrix $\mathbf{A}$ with $\tilde{\mathbf{A}}$. In the case of image reconstruction, low-rank matrix approximation will discard small details in the image, and the amount of the loss of details is inversely proportional to $n$. \figref{fig:LowrankMatrixApproximation} shows an example of the low-rank matrix approximation of an image. Note that the amount of preserved image detail is proportional to the number of preserved singular values.


We extract a SVD feature based on low-rank matrix approximation. Because more non-zero singular values are needed to preserve the details in an image, a sharp image patch tends to have more non-zero singular values than a blurry image patch; i.e., a non-zero $\lambda_{k}$ with large $k$ is a clue to measure the amount of detail. The scaled singular values define our last hand-crafted feature as follows:
\begin{equation}
f_{S}(k) = \frac{1}{W_{S}}\log\left(1+\lambda_{k}\right), \ k \in [1, n_{S}],
\label{eq:fSVD}
\end{equation}
where $W_{S}$ denotes the normalization factor and $n_{S}$ denotes the dimensions of the feature. \figref{fig:HandCraftedFeatures} (c) shows a comparison of sharp and blurry SVD features. The long tail of the sharp feature implies that more details are preserved in an image patch.

\subsection{Deep Feature}
We extract the deep feature $f_{C}$ from a color image patch using a CNN. To deal with multi-scale patches, small-scale patches are zero-padded before they are fed into the CNN. Our feature extraction network consists of convolutional, ReLU and max pooling layers, as illustrated in \figref{fig:CNN}. Successive convolutional, ReLU and max pooling layers are suitable to obtain highly non-linear features. Because the deep feature is learnt from a massive amount of training data (\secref{subsec:ClassifierTraining}), it can accurately
distinguish between sharp and blurry features. In addition, it compensates for the lack of color and cross-channel information in the hand-crafted features, which are important and valuable for our task. \figref{fig:CNN} also shows the average outputs from our feature extraction network with sharp, intermediate and blurry image patches. The activations are proportional to the sharpness of the input image.

\subsection{Our Defocus Feature}
\label{subsec:defocusfeature}

\begin{table}[t]
\begin{center}
\def\arraystretch{1.2}
\begin{tabular}{|M{0.20}|M{0.18}||M{0.20}|M{0.18}|}
\hline
{\scriptsize Feature} & {\scriptsize Accuracy(\%)} & {\scriptsize Feature} & {\scriptsize Accuracy(\%)} \\ \hline
$f_{D}$ {\tiny (DCT)} & 38.15 & $f_{H}$ {\tiny (Hand-crafted)} & 71.49 \\ \hline
$f_{G}$ {\tiny (Gradient)} & 68.36 & $f_{C}$ {\tiny (Deep)} & 89.38 \\ \hline
$f_{S}$ {\tiny (SVD)} & 61.76 & $f_{B}$ {\tiny (Concatenated)} & 94.16 \\ \hline
\end{tabular}
\end{center}
\vspace{-0.2in}
\caption{Classification accuracies. Note that the accuracy of a random guess is 9.09 \%.}
\label{table:FeatureAccuracy}
\vspace{-0.2in}
\end{table}

We concatenate all of the extracted features to construct our final defocus feature $f_{B}$ as follows:
\begin{equation}
f_{B} = [f_{H}, \ f_{C}] = [[f_{D}, \ f_{G}, \ f_{S}], \ f_{C}],
\label{eq:FeatureConcatenation}
\end{equation}
where $[\cdot]$ denotes the concatenation. Table \ref{table:FeatureAccuracy} shows the classification accuracy of each feature. We use a neural network classifier for the accuracy comparison. The classification tolerance is set to an absolute difference of 0.15 compared to the standard deviation value $\sigma$ of the ground truth blur kernel. We train neural networks with the same architecture using those features individually and test on 576,000 features of 11 classes. The details of the classifier and the training process will be presented in Sections \ref{subsec:Classifier} and \ref{subsec:ClassifierTraining}. Our deep feature, $f_{C}$, has the most discriminative power with regard to blurry and sharp features as compared to other individual hand-crafted features, $\{f_{D}, f_{G}, f_{S}\}$, and their concatenation, $f_{H}$. When all hand-crafted and deep features are concatenated ($f_{B}$), the performance is even more enhanced. Removing one of the hand-crafted features drops the performance by approximately 1-3\%. For example, the classification accuracies of $[f_{D}, f_{S}, f_{C}]$ and $[f_{G}, f_{S}, f_{C}]$ are 93.25\% and 91.10\%, respectively. In addition, the performance of $f_{B}$ with only single-scale patch extraction also decreases to 91.00\%.

\section{Defocus Map Estimation}
\label{sec:DefocusMapEstimation}
The defocus features are classified using a neural network classifier to determine the amount of defocus at the center point of each patch. We obtain an initial sparse defocus map after the classification step and then filter out a number of outliers from the sparse defocus map using a sparse joint bilateral filter~\cite{zhuo2011defocus} adjusted with the classification confidence values. A full defocus map is estimated from the filtered sparse defocus map using a matting Laplacian algorithm~\cite{levin2008closed}. The edge-preserving smoothing filtered~\cite{zhang2014rolling} color image is used as a guidance of propagation.

\subsection{Neural Network Classifier}
\label{subsec:Classifier}
We adopt a neural network classifier for classification because it can capture the highly non-linear relationship between our defocus feature components and the amount of defocus. Moreover, its outstanding performance has been demonstrated in various works~\cite{karpathy2014large, ILSVRC15, simonyan2014very}. Our classifier network consists of three fully connected layers (300-150-11 neurons each) with ReLU and dropout layers. The softmax classifier is used for the last layer. Details about the classifier training process will be presented in Sections \ref{subsec:ClassifierTraining} and \ref{subsec:ScaleEncoding}. Using this classification network, we obtain the labels and probabilities of features, after which the labels are converted to the corresponding $\sigma$ values of the Gaussian kernel, which describe the amount of defocus. Subsequently, we construct the sparse defocus map $I_{S}$ using the $\sigma$ values and the confidence map $I_{C}$ using the probabilities.

\subsection{Sparse Defocus Map}
\begin{figure}
\begin{center}
\begin{tabular}{@{}c}
\includegraphics[width=0.990\linewidth]{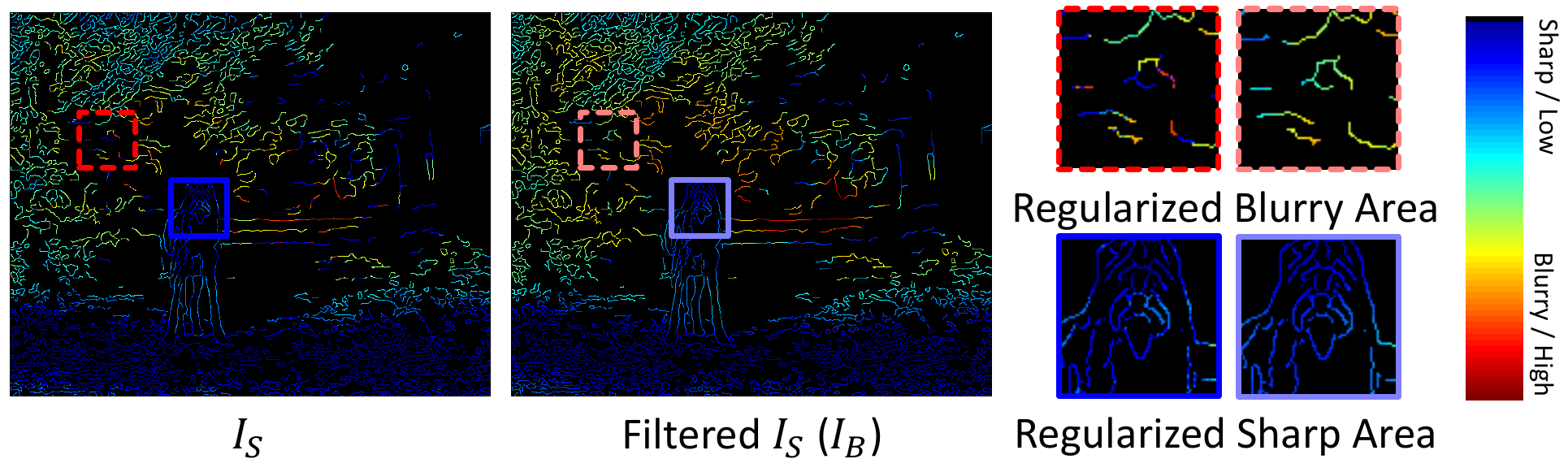}
\end{tabular}
\end{center}
\vspace{-0.25in}
\caption{Sparse defocus maps before ($I_{S}$) and after ($I_{B}$) probability-joint bilateral filtering. Solid boxes denote sharp areas (small value) and dashed boxes denote blurry areas (large value). Some outliers are filtered out effectively.}
\label{fig:ClassifierAndJBF}
\vspace{-0.2in}
\end{figure}

The sparse defocus map is filtered by the probability-joint bilateral filter to reject certain outliers and noise. In addition, we filter the input image with an edge-preserving smoothing filter to create a guidance image $I_{G}$ for probability-joint bilateral filtering and sparse defocus map propagation. A rolling guidance filter~\cite{zhang2014rolling} is chosen because it can effectively remove image noise together with the distracting small-scale image features and prevent erroneous guidances on some textured and noisy regions. Our probability-joint bilateral filter $B(\cdot)$ is defined as follows:
\begin{multline}
B(\mathbf{x}) = \frac{1}{W(\mathbf{x})}\sum_{\mathbf{p} \in \mathcal{N}_{\mathbf{x}}}G_{\sigma_{s}}(\mathbf{x},\mathbf{p}) \\
\times G_{\sigma_{r}}(I_{G} (\mathbf{x}),I_{G}(\mathbf{p}))G_{\sigma_{c}}(1,I_{C}(\mathbf{p}))I_{S}(\mathbf{p}),
\label{eq:JBF}
\end{multline}
where $G_{\sigma}(\mathbf{u}, \mathbf{v}) = \mathrm{exp}(-\Arrowvert \mathbf{u} - \mathbf{v} \Arrowvert_{F}^{2} / 2\sigma^{2})$  and $\Arrowvert \cdot \Arrowvert_{F}$, $W$, $\mathcal{N}_{\mathbf{x}}$, $\sigma_{s}$, $\sigma_{r}$ and $\sigma_{c}$ denote the Frobenius norm, the normalization factor, the non-zero neighborhoods of $\mathbf{x}$, the standard deviation of spatial, range and probability weight, respectively. The probability weight $G_{\sigma_{c}}$ is added to the conventional joint bilateral filter in order to reduce the unwanted effects of the outliers within neighborhood with low probability values. Probability-joint bilateral filtering removes outliers and regularizes the sparse defocus map effectively as shown in \figref{fig:ClassifierAndJBF}.

\subsection{Full Defocus Map}
The full defocus map is obtained from the sparse defocus map using the matting Laplacian algorithm~\cite{levin2008closed} with the help of the guidance image $I_{G}$. The matting Laplacian is defined as follows:
\begin{multline}
L(i, j) = \sum_{k|(i,j) \in \mathrm{w}_{k}}\Big(\delta_{ij} - \frac{1}{|\mathrm{w}_{k}|}\big(1 + (I_{G}(i) - \mu_{k}) \\
\times (\Sigma_{k} + \frac{\epsilon}{|\mathrm{w}_{k}|}I_{3})^{-1}(I_{G}(j) - \mu_{k})\big)\Big),
\label{eq:MattingLaplacian}
\end{multline}
where $i$, $j$, $k$, $\mathrm{w}_{k}$, $|\mathrm{w}_{k}|$, $\delta_{ij}$, $\mu_{k}$, $\Sigma_{k}$, $I_{3}$ and $\epsilon$ denote the linear indices of pixels, a small window centered at $k$, the size of $\mathrm{w}_{k}$, the Kronecker delta, the mean and variance of the $\mathrm{w}_{k}$, a 3$\times$3 identity matrix and a regularization parameter, respectively. The full defocus map $I_{F}$ is obtained by solving the following least-squares problem:
\begin{equation}
\hat{I}_{F} = \gamma\ (L + D_{\gamma})^{-1}\ \hat{I}_{B},
\label{eq:LeastSquaresProblem}
\end{equation}
where $\hat{I}$, $\gamma$ and $D_{\gamma}$ denote the vector form of matrix $I$, a user parameter and a diagonal matrix whose element $D(i, i)$ is $\gamma$ when $I_B(i)$ is not zero, respectively. An example of a full defocus map is shown in \figref{fig:Teaser} (b).

\section{Experiments}
\label{sec:Experiments}
We first describe our method to generate the training data and train the classifier, after which we compare the performance of our algorithm with the performances of state-of-the-art algorithms using a blur detection dataset~\cite{shi2014discriminative}. Various applications such as blur magnification, all-in-focus image generation and 3-D estimation are also presented. Our codes and dataset are available on our project page.\footnote{\textcolor{magenta}{\url{https://github.com/zzangjinsun/DHDE_CVPR17}}}



\subsection{Classifier Training}
\label{subsec:ClassifierTraining}
For the classification network training, 300 sharp images are randomly selected from the ILSVRC~\cite{ILSVRC15} training data. Similar to the feature extraction step, we extract approximately 1M multi-scale image patches on strong edges and regard these patches as sharp patches. After that, each sharp patch $P_{I}^{S}$ is convolved with synthetic blur kernels to generate blurry patches $P_{I}^{B}$ as follows:
\begin{equation}
P_{I}^{B_{l}} = P_{I}^{S} \ast h_{\sigma_{l}}\ ,\ l \in [1, L],
\label{eq:SynPatchBlur}
\end{equation}
where $h_{\sigma}$, $\ast$ and $L$ denote the Gaussian blur kernel with a zero mean and variance $\sigma^{2}$, the convolution operator and the number of labels, respectively. We set $L = 11$ and the $\sigma$ values for each blur kernel are then calculated as follows:
\begin{equation}
\sigma_{l} = \sigma_{min} + (l-1)\sigma_{inter},
\label{eq:SigmaCalculation}
\end{equation}
where we set $\sigma_{min} = 0.5$ and $\sigma_{inter} = 0.15$.

For the training of the deep feature, $f_{C}$, we directly connect the feature extraction network and the classifier network to train the deep feature and classifier simultaneously. The same method is applied when we use the concatenated feature, $f_{H}$, for training. For the training of $f_{B}$, we initially train the classifier connected to the feature extraction network only (i.e., with $f_{C}$ only), after which we fine-tune the classifier with the hand-crafted features, $f_{H}$. We use the Caffe~\cite{jia2014caffe} library for our network implementation.

\begin{figure}
\begin{center}
\begin{tabular}{@{}c}
\includegraphics[width=0.950\linewidth]{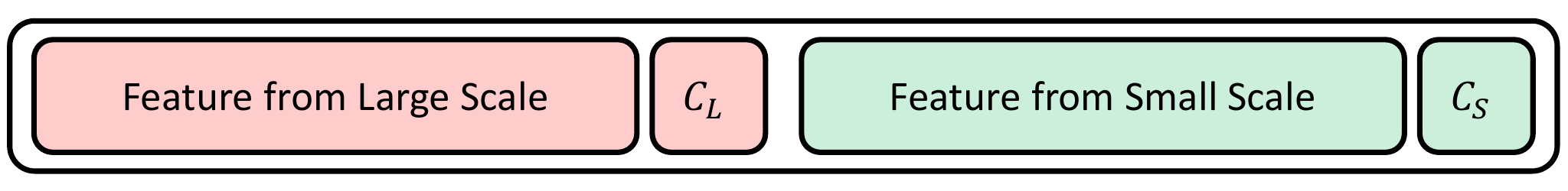}
\end{tabular}
\end{center}
\vspace{-0.3in}
\caption[Our feature scale encoding scheme.]{Our feature scale encoding scheme. If a feature is from large-scale, we fill the small-scale positions with zeros, and vice versa.}
\label{fig:ScaleEncoding}
\vspace{-0.2in}
\end{figure}

\subsection{Scale Encoding}
\label{subsec:ScaleEncoding}
While we trained the classifier with features from small and large patches together, we found that the network does not converge. This occurs because some features from different scales with different labels can be similar. If we train classifiers for each scale individually, it is inefficient and risky, as the parameters to be trained are doubled and there is no assurance that individual networks are consistent.

In order to encode scale information to a feature itself, we assign specific positions for features from each scale, as shown in \figref{fig:ScaleEncoding}. An entire feature consists of positions and constants for each scale. Additional constants are assigned to deal with biases for each scale.

In neural networks, the update rules for weights and the bias in a fully connected layer are as follows:
\begin{equation}
w_{jk}^{t} \rightarrow w_{jk}^{t} - \eta(a_{k}^{t-1}\delta_{j}^{t}),
\label{eq:UpdateWeights}
\end{equation}
\begin{equation}
b_{j}^{t} \rightarrow b_{j}^{t} - \eta\delta_{j}^{t},
\label{eq:UpdateBias}
\end{equation}
where $w_{jk}^{t}$, $b_{j}^{t}$, $\eta$, $a$ and $\delta$ denote the $k$-th weight in the $j$-th neuron of the layer $t$, the bias of the $j$-th neuron of the layer $t$, the learning rate, the activation, and the error, respectively. For the input layer $(t=1)$, if we set the $k$-th dimension of the input feature to a constant $C$, \eqnref{eq:UpdateWeights} becomes $w_{jk}^{1} \rightarrow w_{jk}^{1} - \eta(C\delta_{j}^{1}),$ and surprisingly, if we let $C = 1$, the update rule takes the same form of the bias update rule. Therefore, by introducing additional constants into each scale, we are able not only effectively to separate the biases for each scale but also to apply different learning rates for each bias with different constants. After encoding scale information to a feature itself, the classifier network converges. In our experiments, we simply set $C_{L} = C_{S} = 1$.

Owing to this encoding scheme, we set the number of neurons in the first layer of the classifier network such that it is roughly two times greater than the dimensions of our descriptor to decode small- and large-scale information from encoded features.

\begin{figure}
\begin{center}
\begin{tabular}{@{}c}
\includegraphics[width=0.990\linewidth]{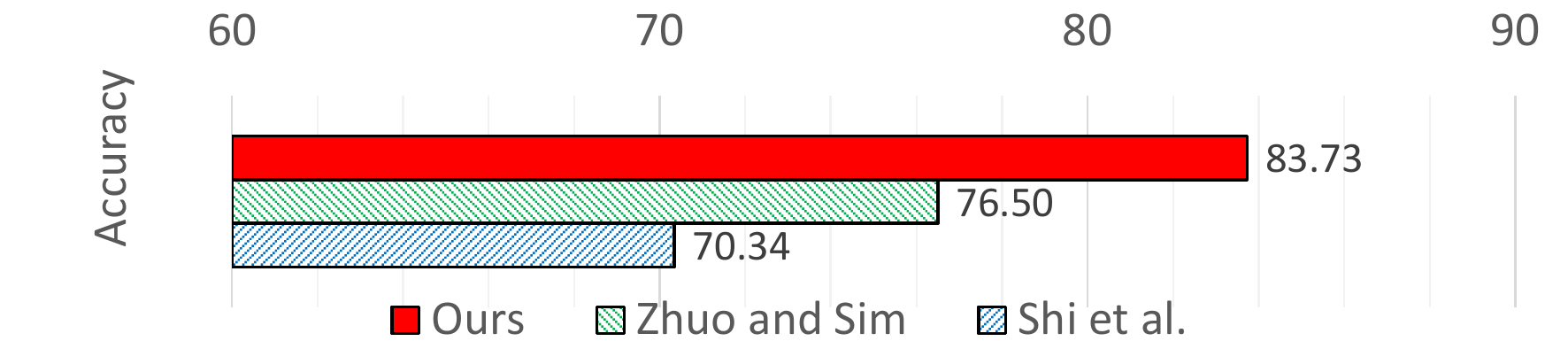} \\
\includegraphics[width=0.990\linewidth]{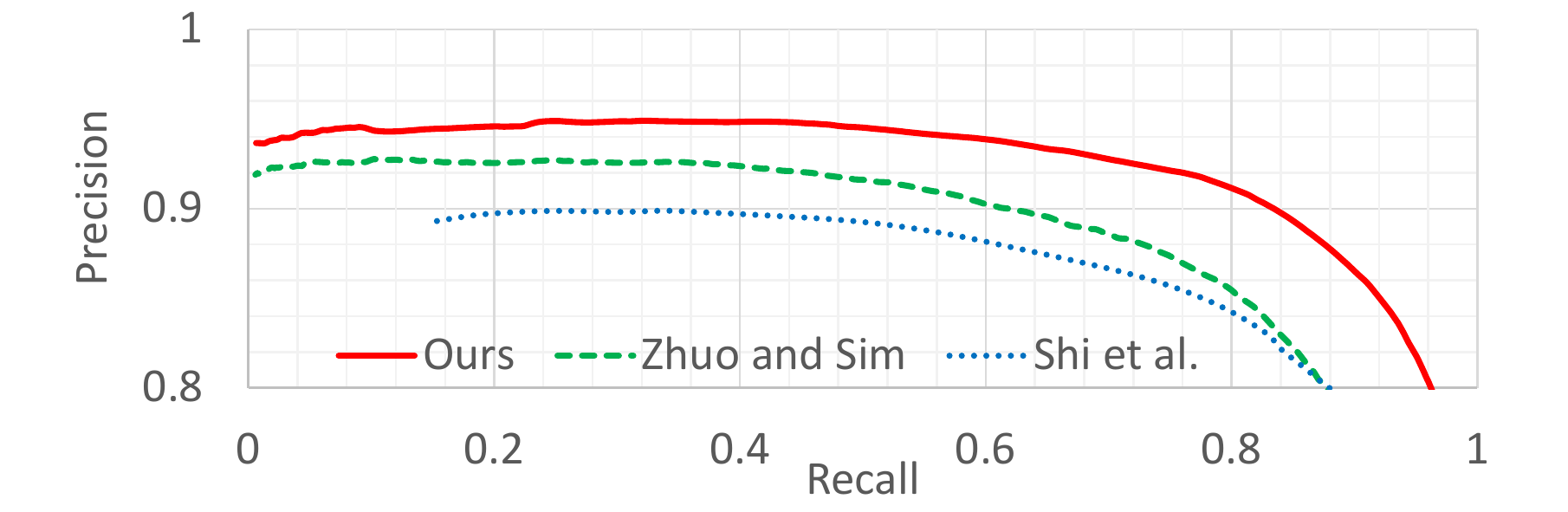}
\end{tabular}
\end{center}
\vspace{-0.3in}
\caption{Segmentation accuracies (top) and Precision-Recall comparison (bottom) of Shi \emph{et al.}~\cite{shi2014discriminative}, Zhuo and Sim~\cite{zhuo2011defocus} and our algorithm.}
\label{fig:QuantitativeComparison}
\vspace{-0.2in}
\end{figure}

\begin{figure*}
\begin{center}
\def\arraystretch{0.5}
\begin{tabular}{@{}c@{\hskip 0.001\linewidth}c@{\hskip 0.001\linewidth}c@{\hskip 0.001\linewidth}c@{\hskip 0.001\linewidth}c@{\hskip 0.001\linewidth}c@{\hskip 0.001\linewidth}c@{\hskip 0.001\linewidth}c}
\includegraphics[width=0.124\linewidth]{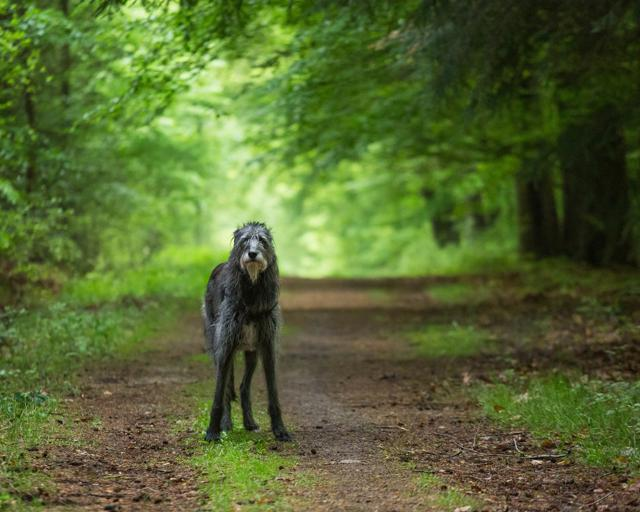} &
\includegraphics[width=0.124\linewidth]{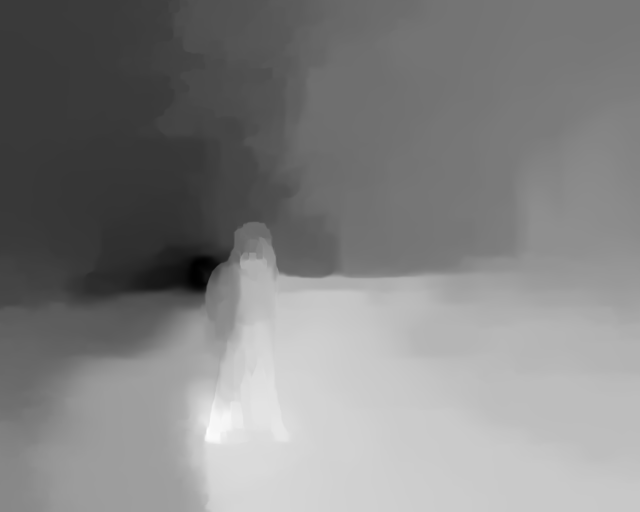} &
\includegraphics[width=0.124\linewidth]{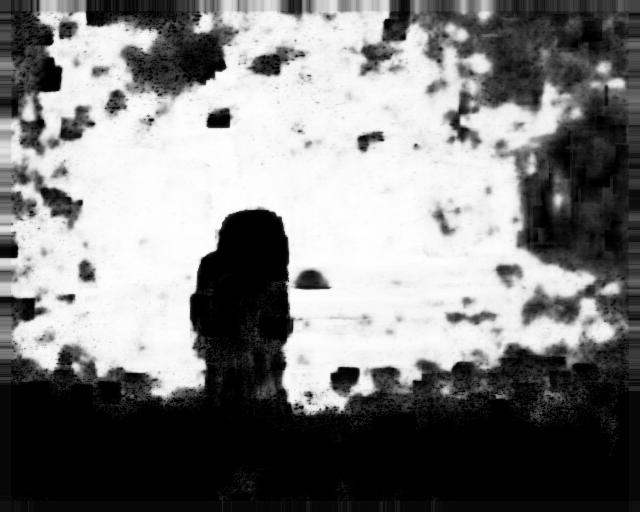} &
\includegraphics[width=0.124\linewidth]{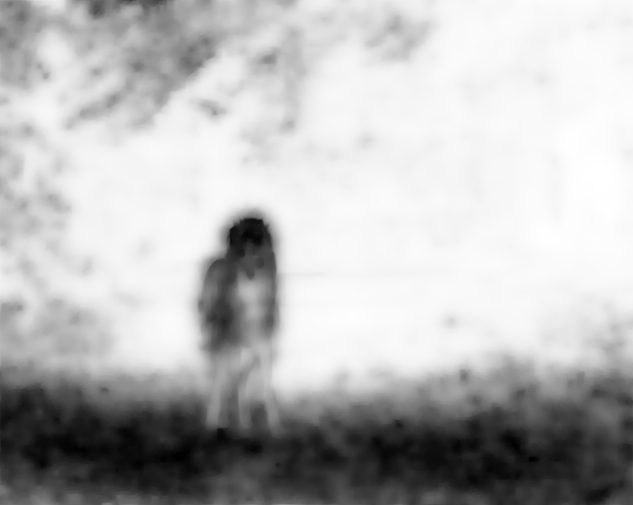} &
\includegraphics[width=0.124\linewidth]{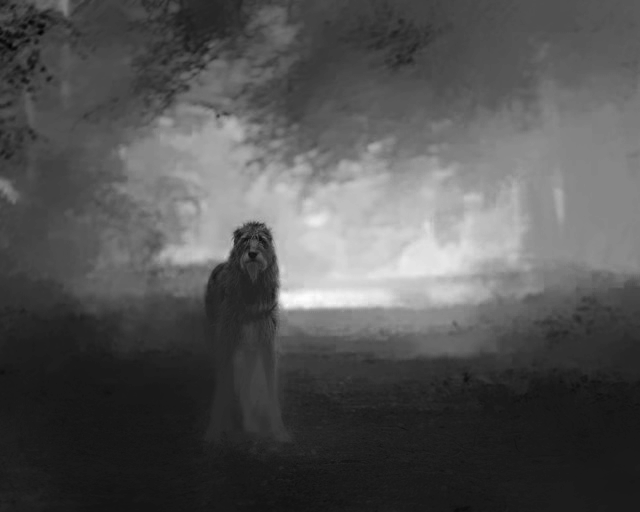} &
\includegraphics[width=0.124\linewidth]{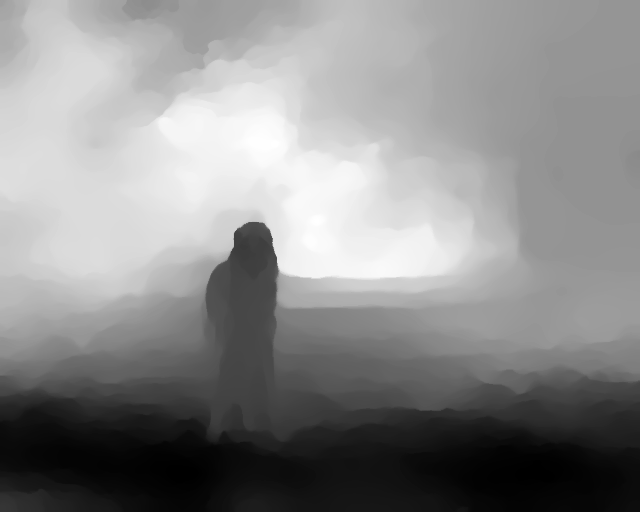} &
\includegraphics[width=0.124\linewidth]{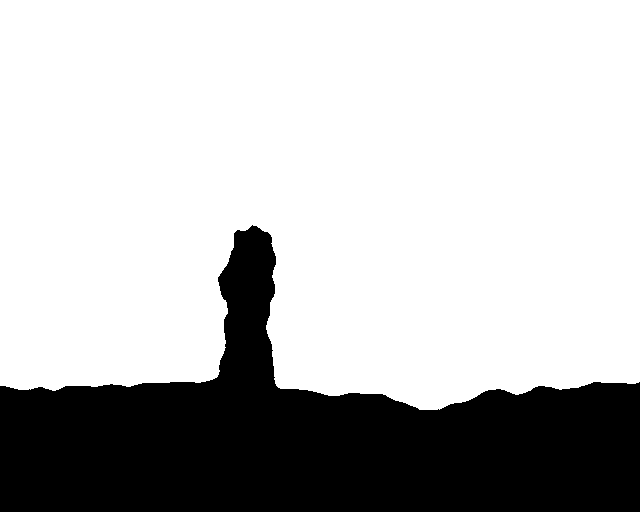} &
\includegraphics[width=0.124\linewidth]{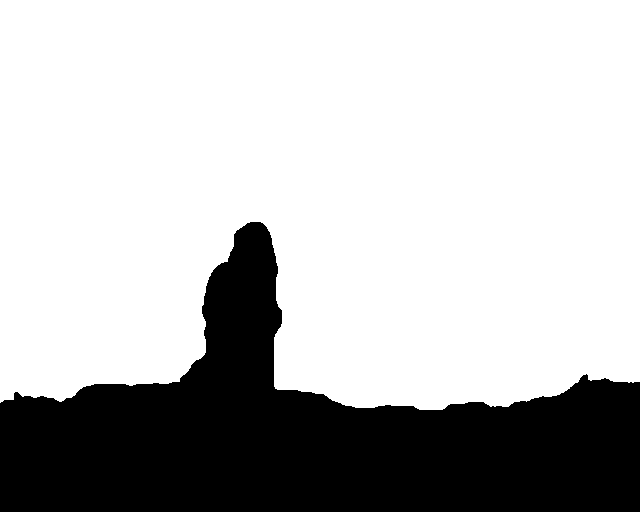} \\
\includegraphics[width=0.124\linewidth]{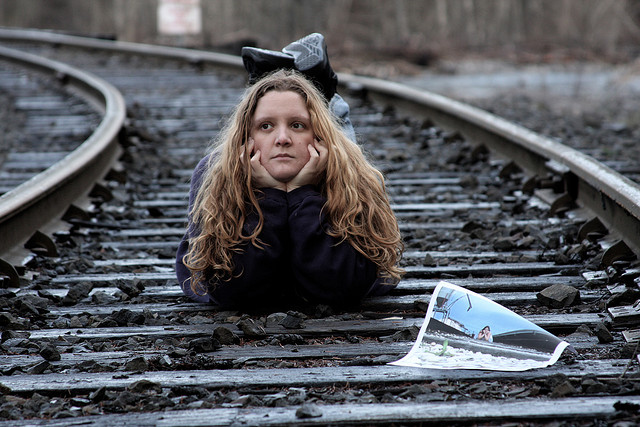} &
\includegraphics[width=0.124\linewidth]{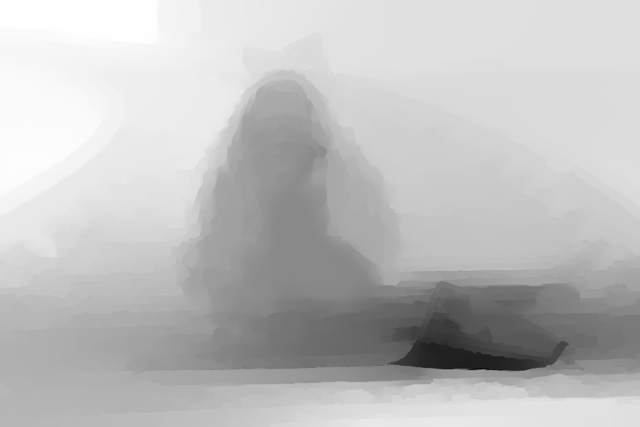} &
\includegraphics[width=0.124\linewidth]{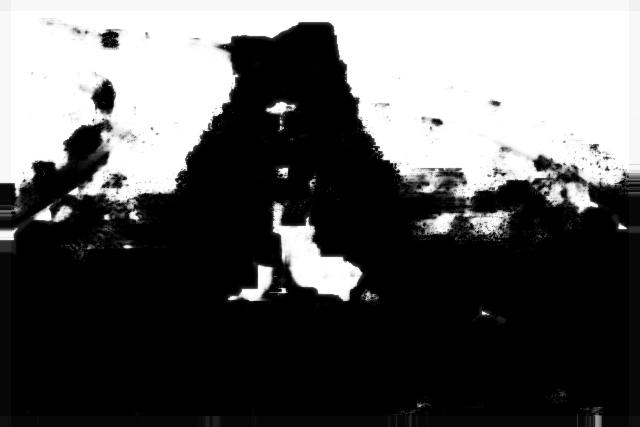} &
\includegraphics[width=0.124\linewidth]{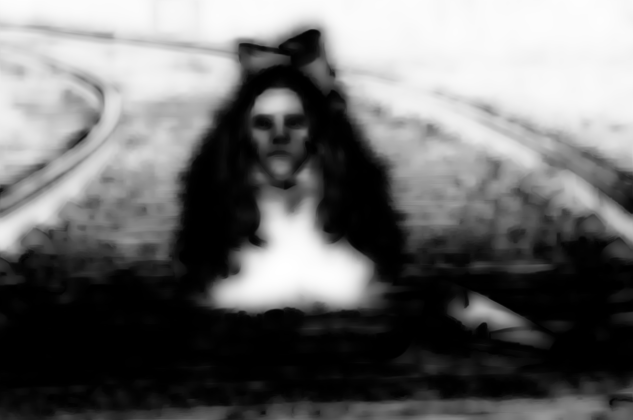} &
\includegraphics[width=0.124\linewidth]{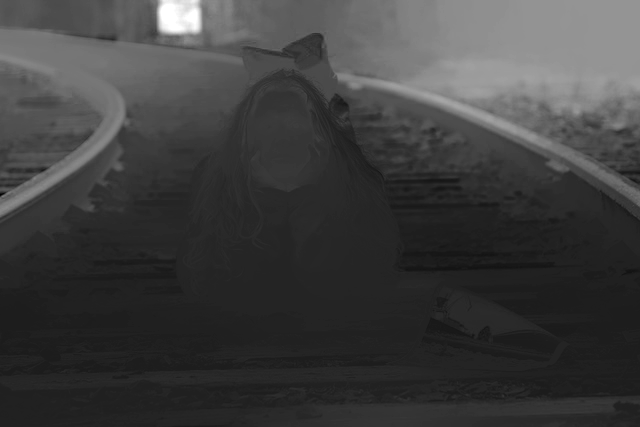} &
\includegraphics[width=0.124\linewidth]{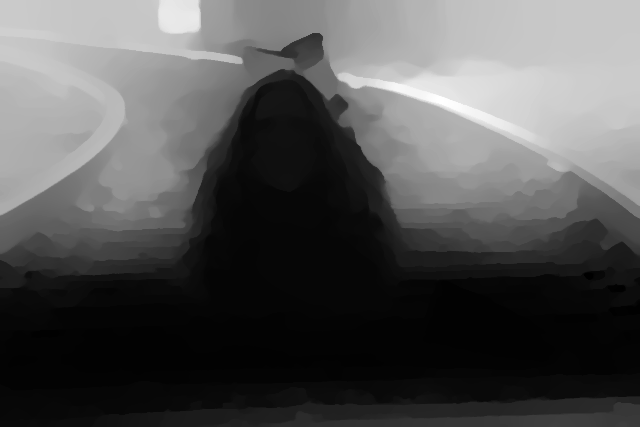} &
\includegraphics[width=0.124\linewidth]{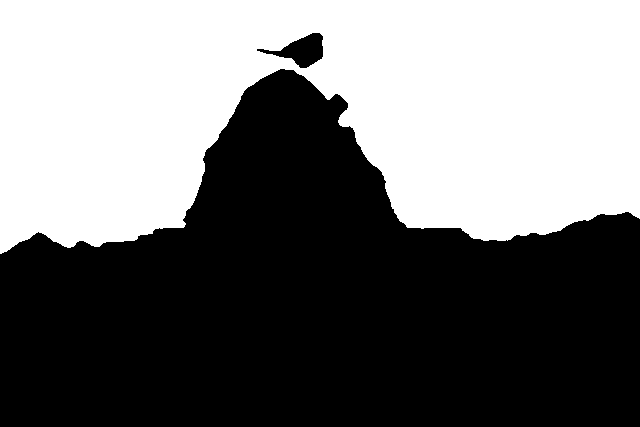} &
\includegraphics[width=0.124\linewidth]{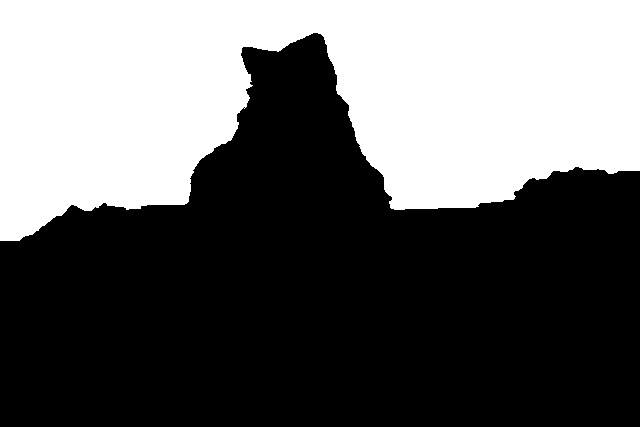} \\
\includegraphics[width=0.124\linewidth]{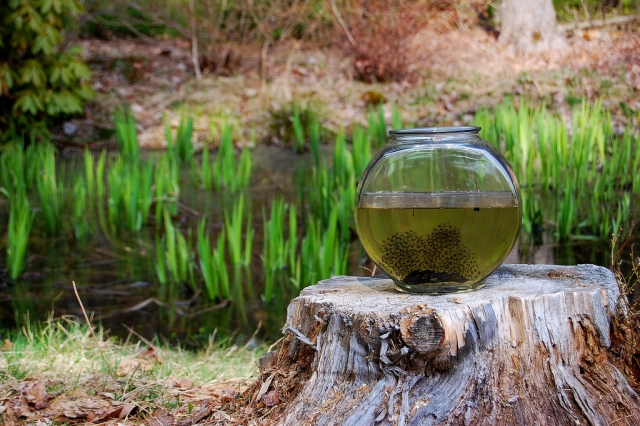} &
\includegraphics[width=0.124\linewidth]{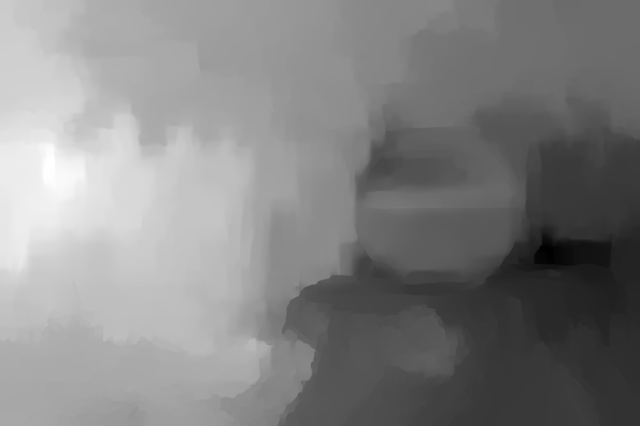} &
\includegraphics[width=0.124\linewidth]{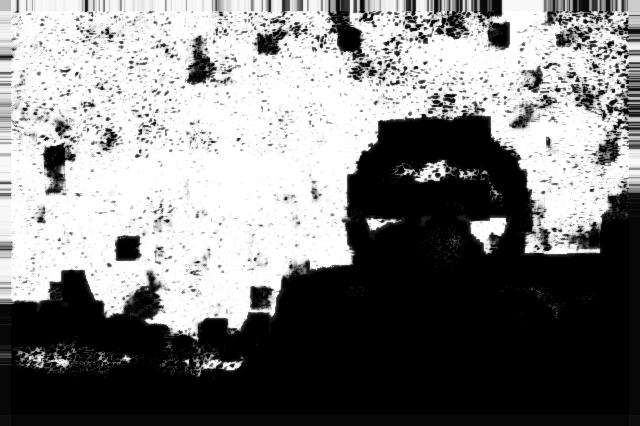} &
\includegraphics[width=0.124\linewidth]{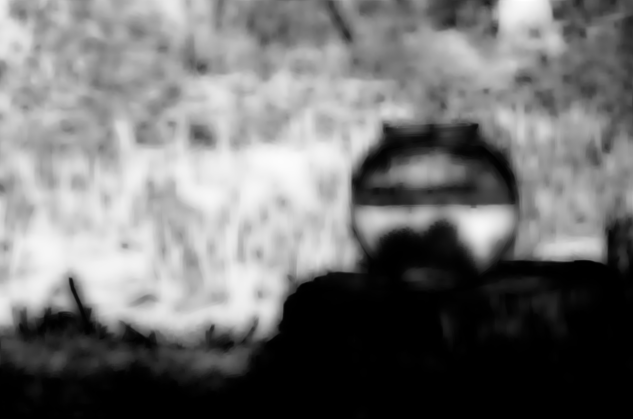} &
\includegraphics[width=0.124\linewidth]{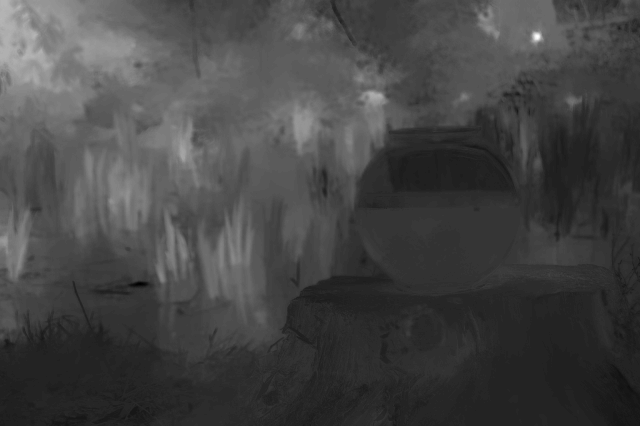} &
\includegraphics[width=0.124\linewidth]{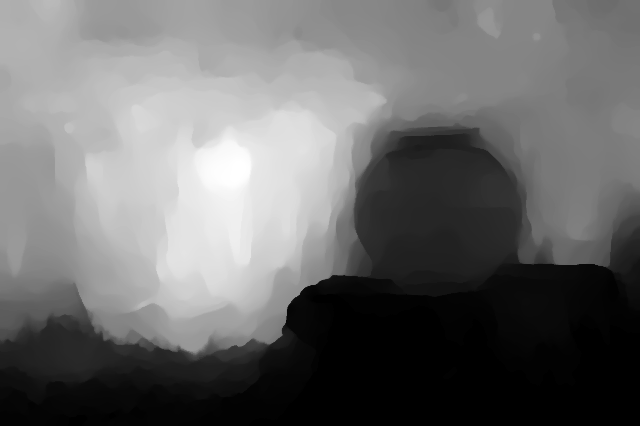} &
\includegraphics[width=0.124\linewidth]{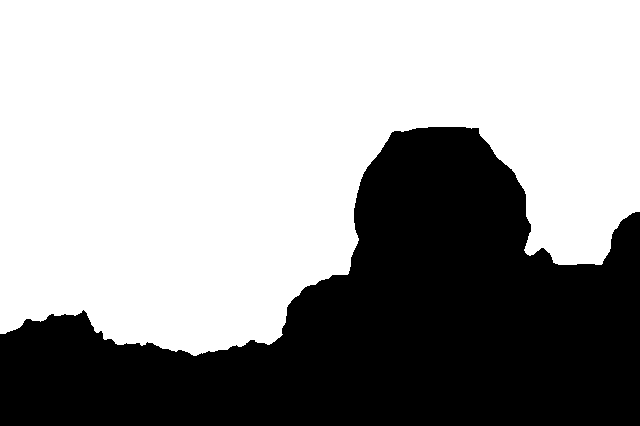} &
\includegraphics[width=0.124\linewidth]{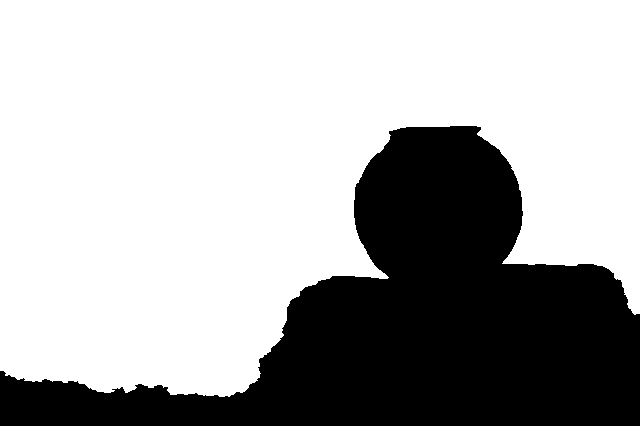} \\
\includegraphics[width=0.124\linewidth]{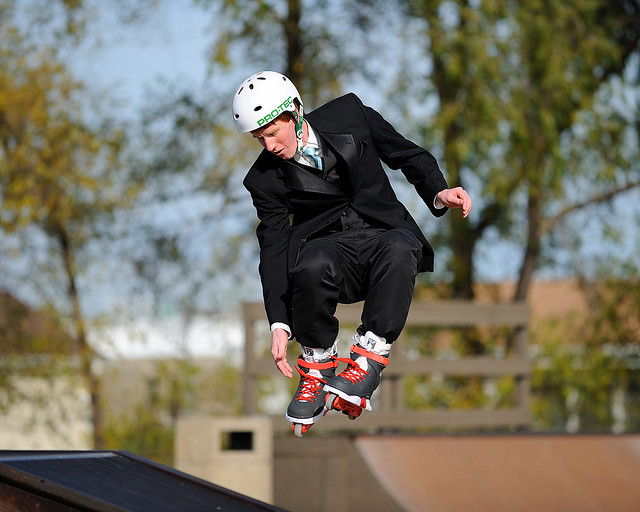} &
\includegraphics[width=0.124\linewidth]{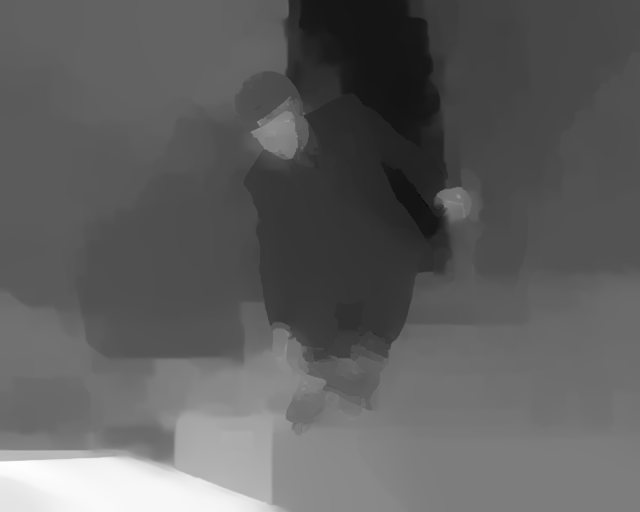} &
\includegraphics[width=0.124\linewidth]{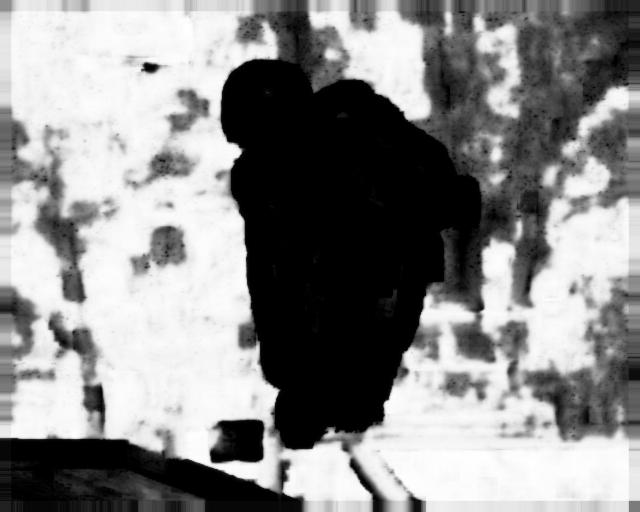} &
\includegraphics[width=0.124\linewidth]{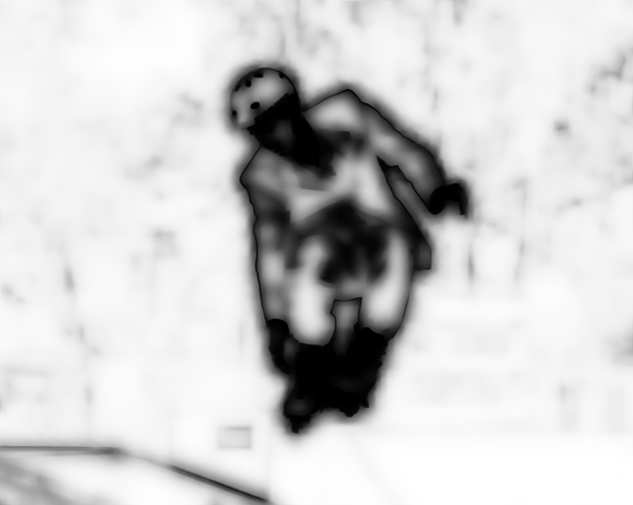} &
\includegraphics[width=0.124\linewidth]{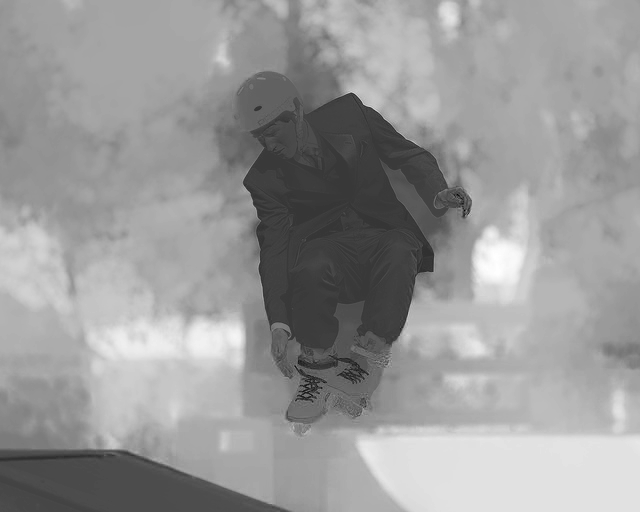} &
\includegraphics[width=0.124\linewidth]{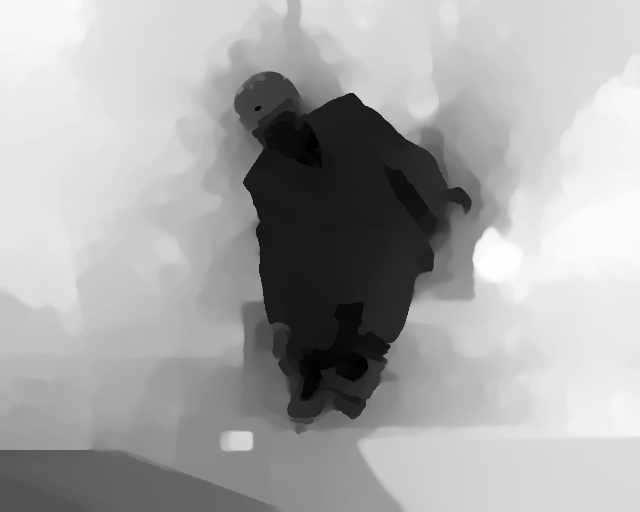} &
\includegraphics[width=0.124\linewidth]{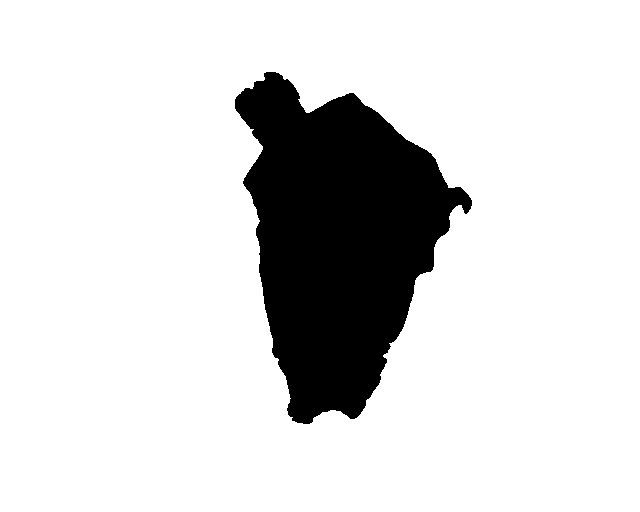} &
\includegraphics[width=0.124\linewidth]{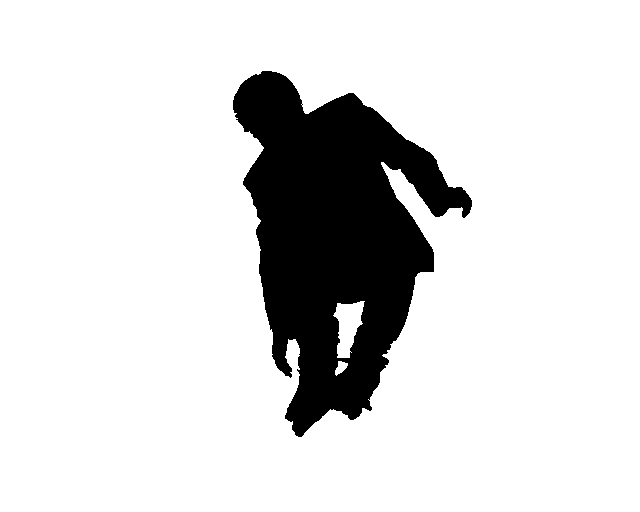} \\
{\small (a)} & {\small (b)} & {\small (c)} & {\small (d)} & {\small (e)} & {\small (f)} & {\small (g)} & {\small (h)}
\end{tabular}
\end{center}
\vspace{-0.25in}
\caption{Defocus map estimation and binary blurry region segmentation results. (a) Input images. (b) Results of~\cite{shi2015break}. (c) Results of~\cite{shi2014discriminative}. (d) Results of~\cite{shi2015just} (Inverted for visualization). (e) Results of~\cite{zhuo2011defocus}. (f) Our defocus maps and (g) corresponding binary masks. (h) Ground truth binary masks.}
\label{fig:Results}
\vspace{-0.25in}

\begin{center}
\def\arraystretch{0.5}
\begin{tabular}{@{}c@{\hskip 0.001\linewidth}c@{\hskip 0.001\linewidth}c@{\hskip 0.001\linewidth}c@{\hskip 0.001\linewidth}c@{\hskip 0.001\linewidth}c@{\hskip 0.001\linewidth}c}
\includegraphics[width=0.142\linewidth]{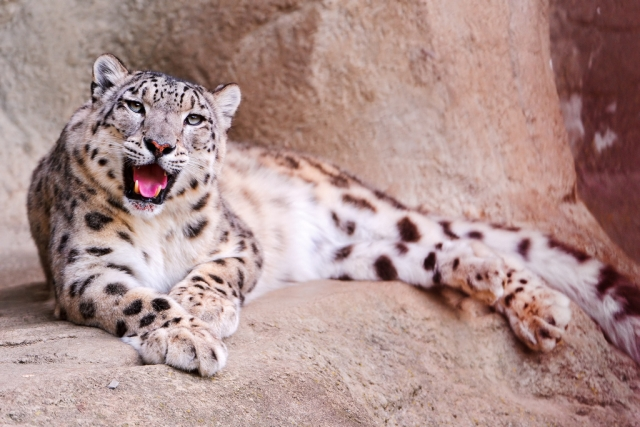} &
\includegraphics[width=0.142\linewidth]{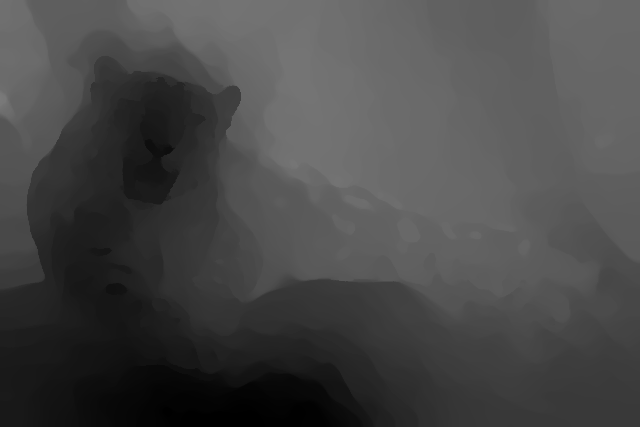} &
\includegraphics[width=0.142\linewidth]{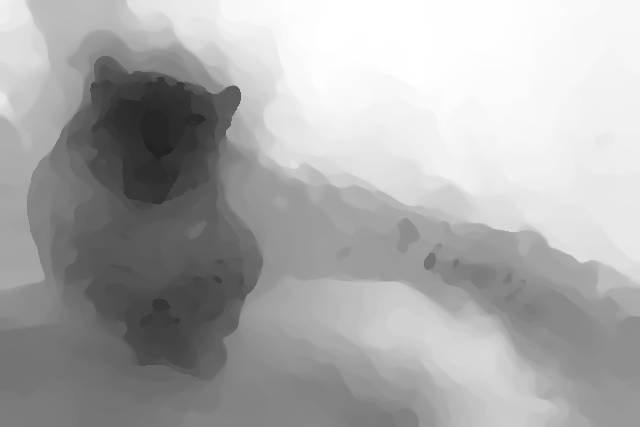} &
\includegraphics[width=0.142\linewidth]{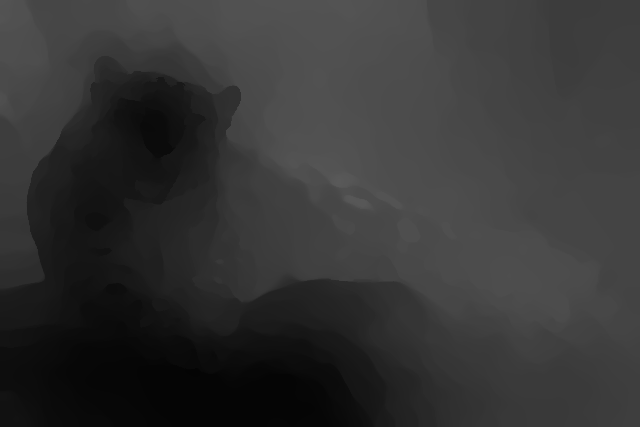} &
\includegraphics[width=0.142\linewidth]{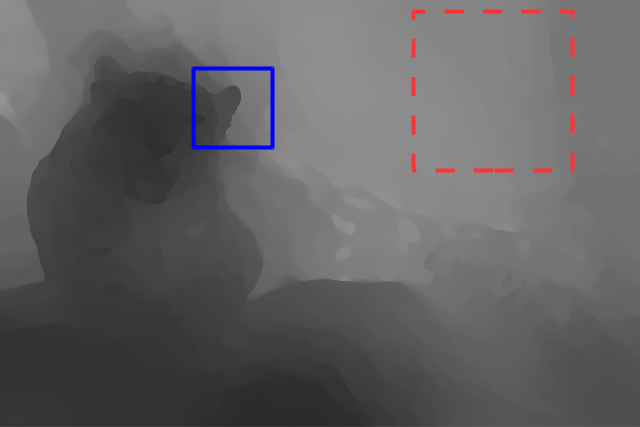} &
\includegraphics[width=0.142\linewidth]{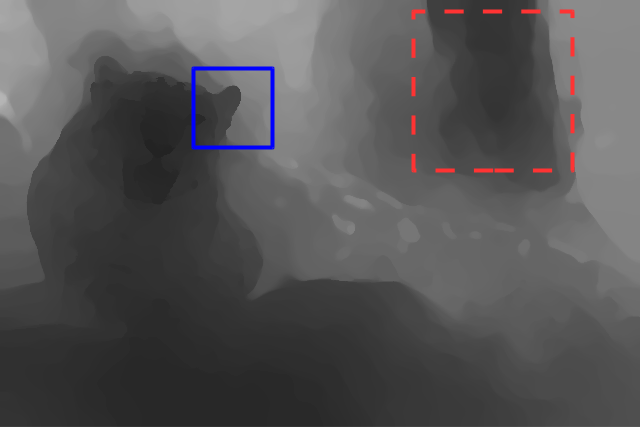} &
\includegraphics[width=0.142\linewidth]{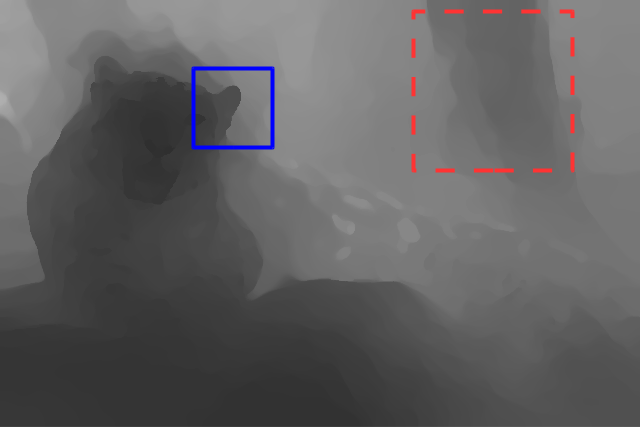} \\
\includegraphics[width=0.142\linewidth]{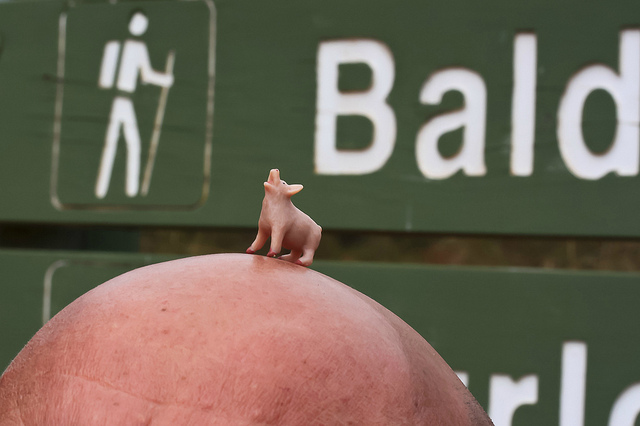} &
\includegraphics[width=0.142\linewidth]{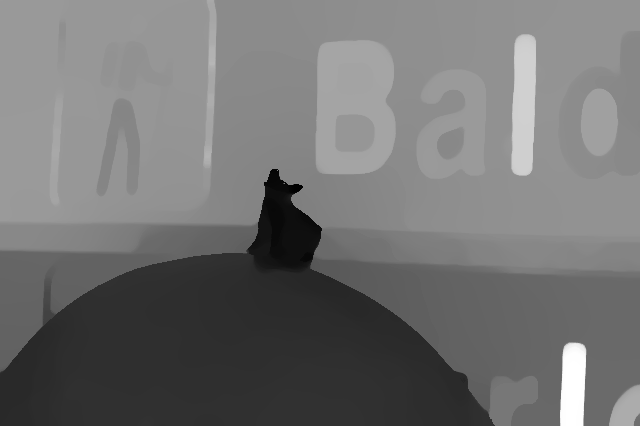} &
\includegraphics[width=0.142\linewidth]{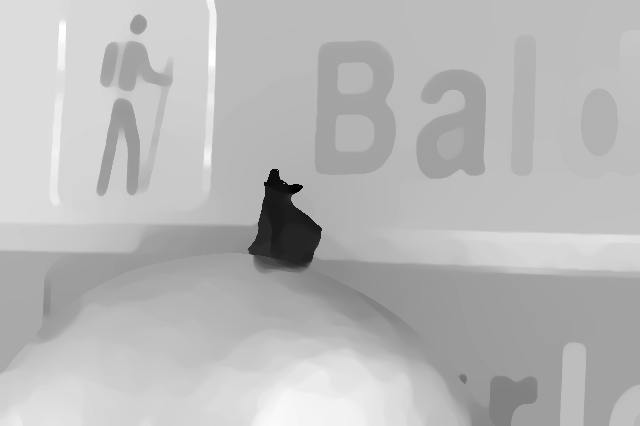} &
\includegraphics[width=0.142\linewidth]{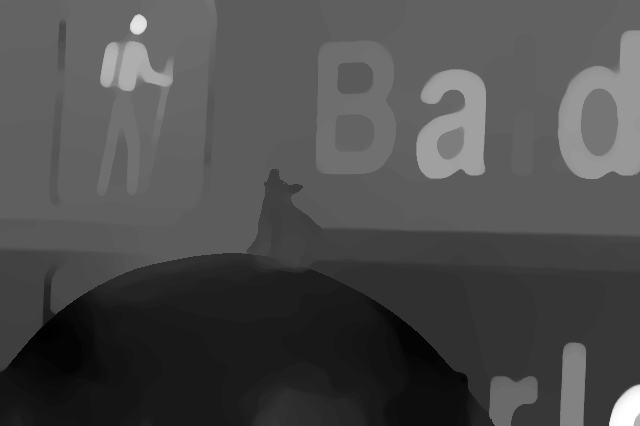} &
\includegraphics[width=0.142\linewidth]{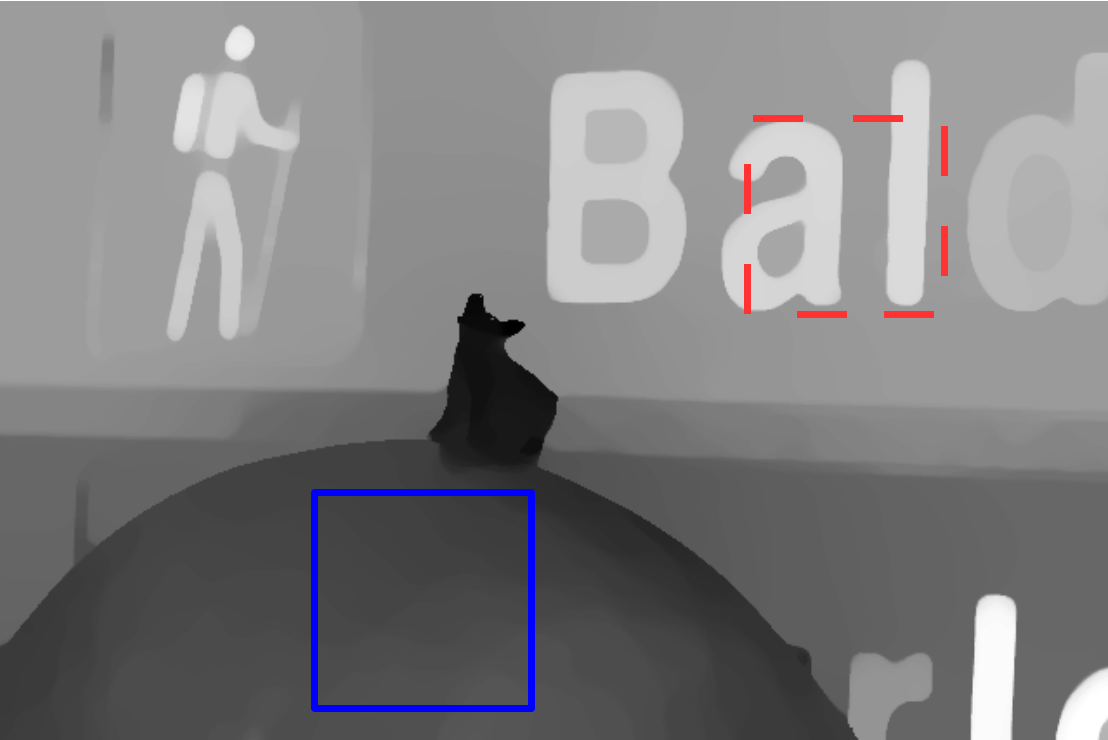} &
\includegraphics[width=0.142\linewidth]{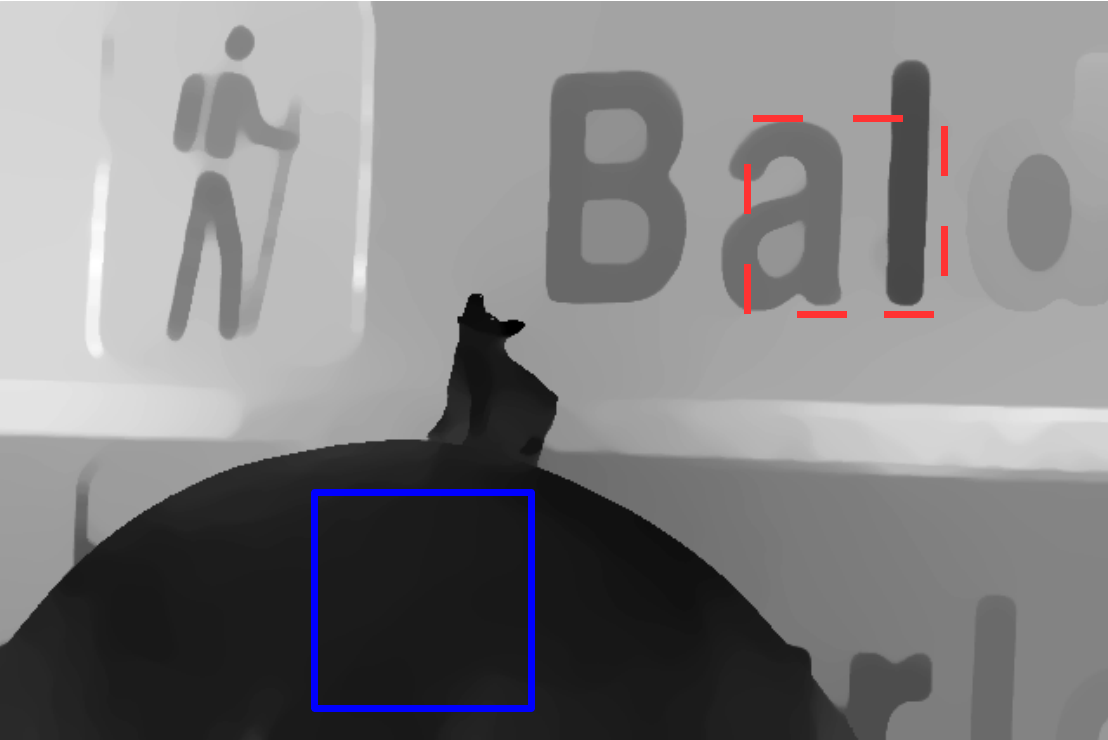} &
\includegraphics[width=0.142\linewidth]{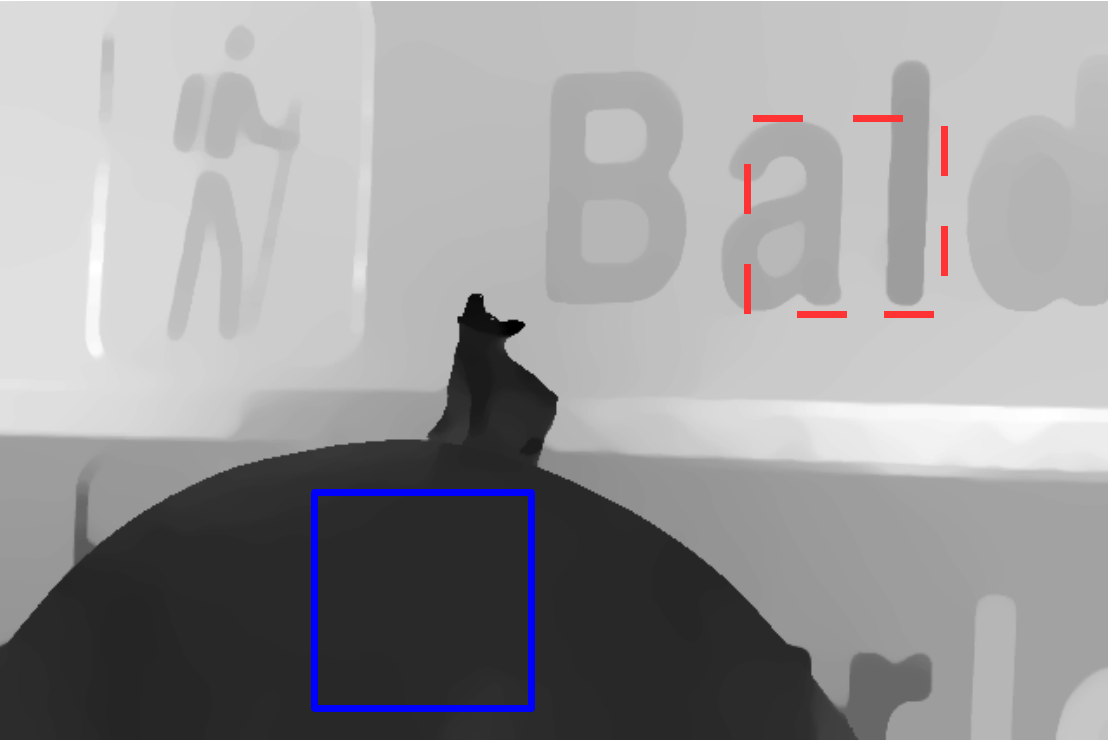} \\
{\small (a) Images} & {\small (b) $f_{D}$} & {\small (c) $f_{G}$} & {\small (d) $f_{S}$} & {\small (e) $f_{C}$} & {\small (f) $f_{H}$} & {\small (g) $f_{B}$}
\end{tabular}
\end{center}
\vspace{-0.25in}
\caption{Defocus maps from each feature. Features used for the estimation of each defocus map are annotated. The blue solid boxes and red dashed boxes in (e), (f) and (g) show the complementary roles of the hand-crafted and deep features.}
\label{fig:ResultWithEachFeature}
\vspace{-0.2in}
\end{figure*} 

\begin{figure}
\begin{center}
\begin{tabular}{@{}c@{\hskip 0.001\linewidth}c@{\hskip 0.001\linewidth}c}
\includegraphics[width=0.332\linewidth]{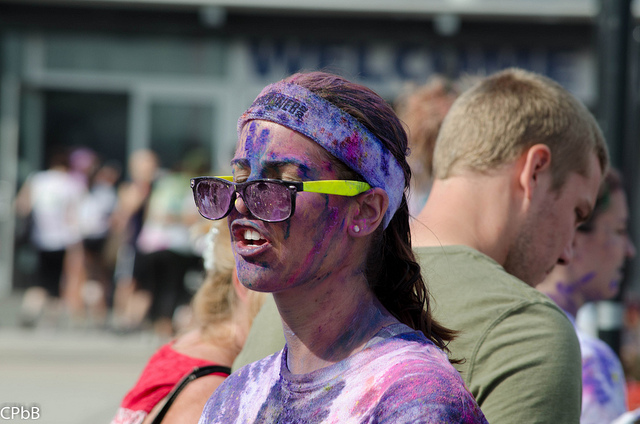} &
\includegraphics[width=0.332\linewidth]{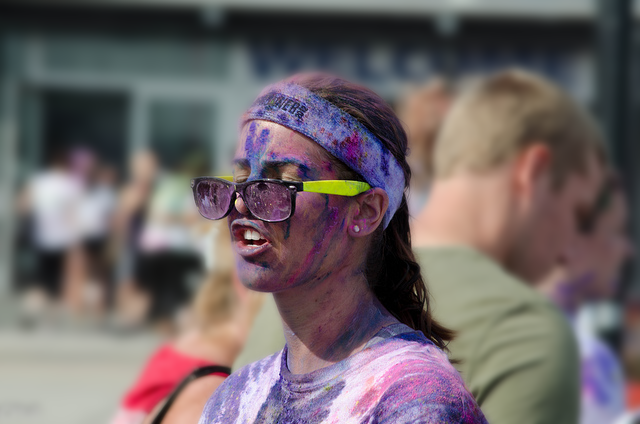} &
\includegraphics[width=0.332\linewidth]{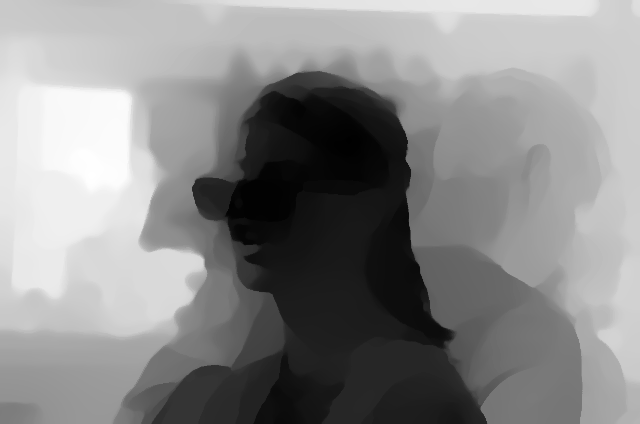} \\
{\small (a) Input} & {\small (b) Magnified} & {\small (c) Defocus map}
\end{tabular}
\end{center}
\vspace{-0.25in}
\caption{Defocus blur magnification. The background blurriness is amplified to direct more attention to the foreground.}
\label{fig:BlurMagnification}
\vspace{-0.25in}

\begin{center}
\begin{tabular}{@{}c@{\hskip 0.001\linewidth}c@{\hskip 0.001\linewidth}c}
\includegraphics[width=0.332\linewidth]{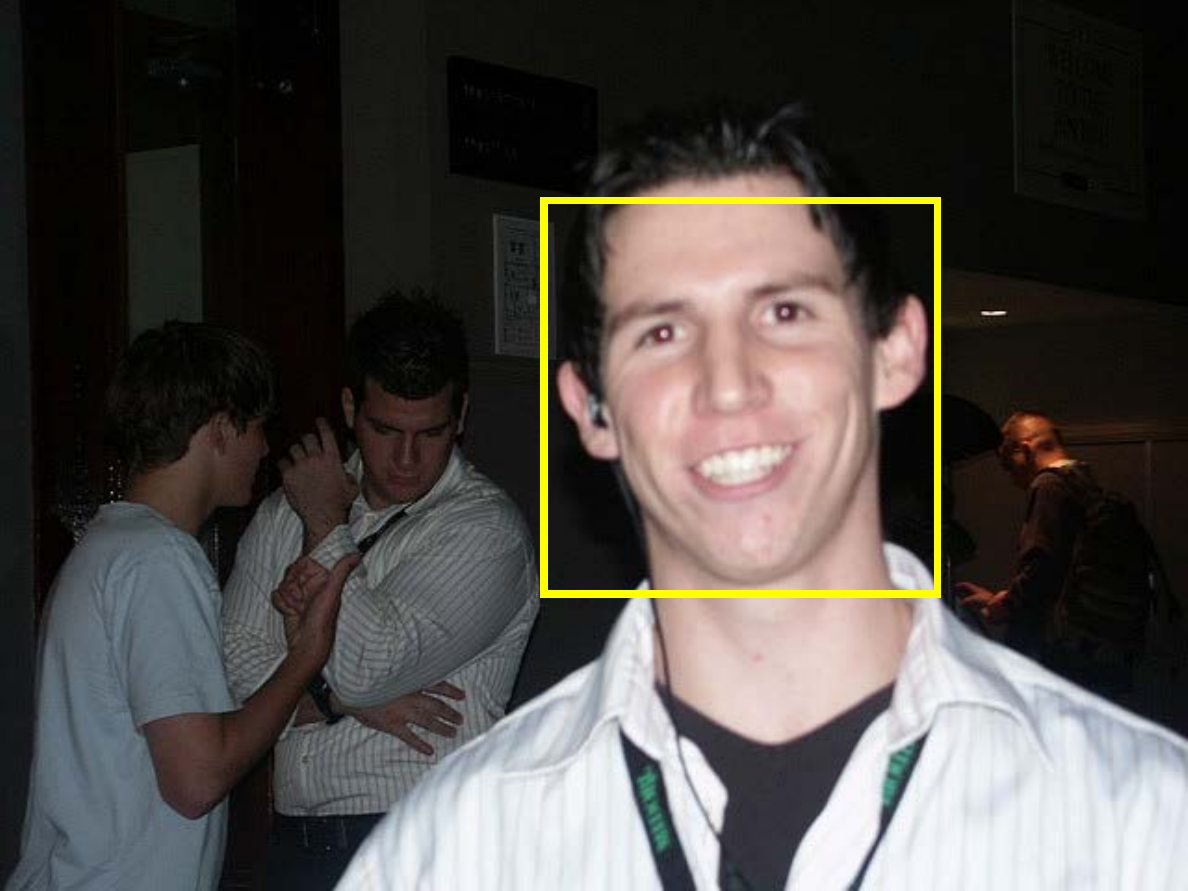} &
\includegraphics[width=0.332\linewidth]{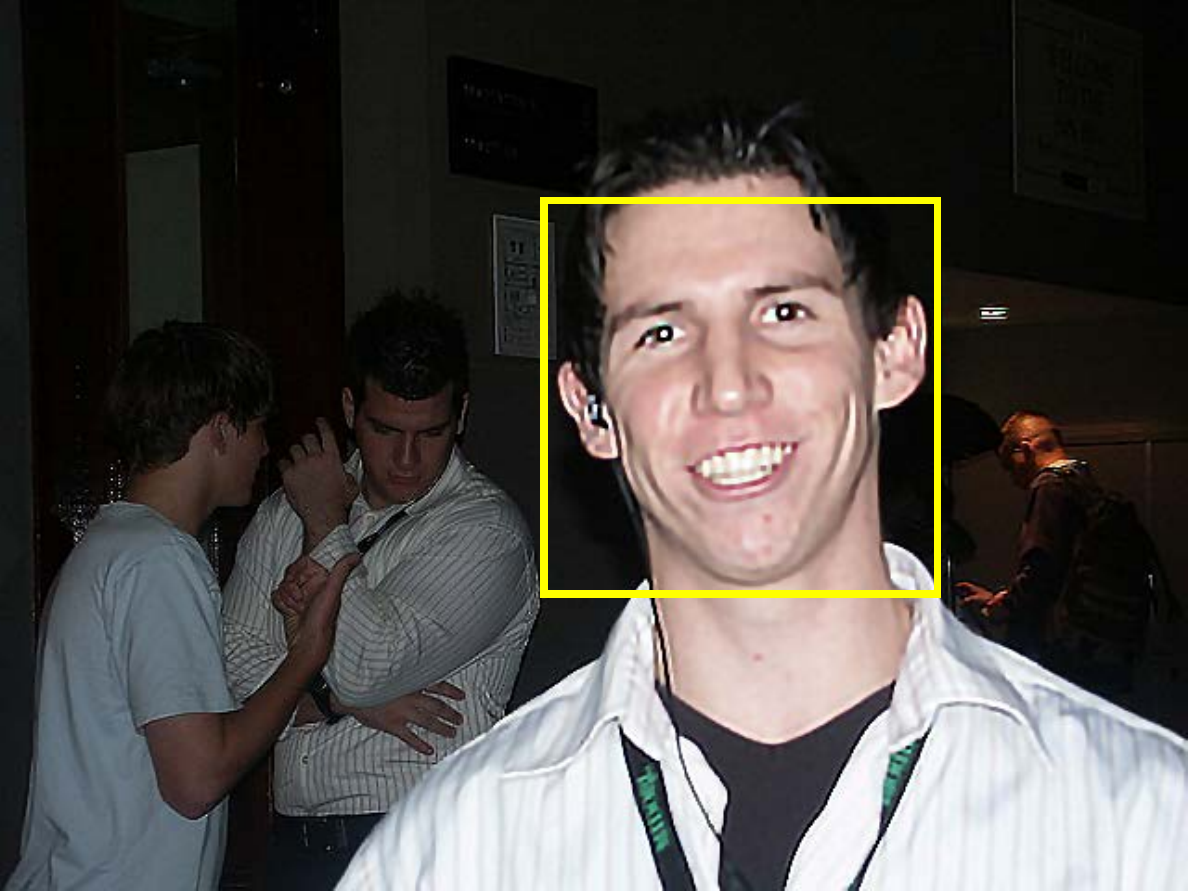} &
\includegraphics[width=0.332\linewidth]{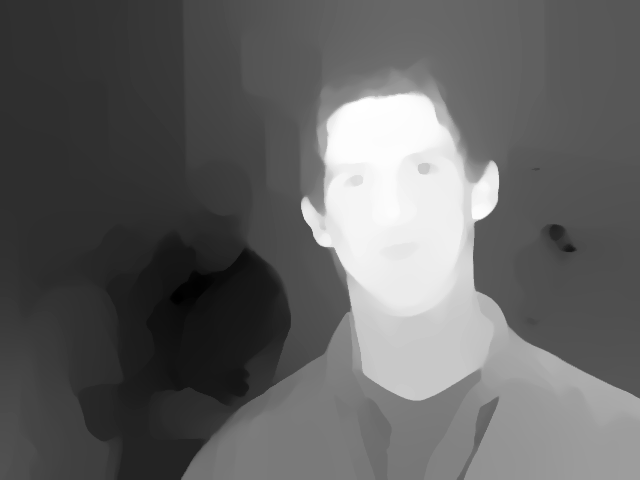} \\
{\small (a) Input} & {\small (b) All-in-focus} & {\small (c) Defocus map}
\end{tabular}
\end{center}
\vspace{-0.25in}
\caption{All-in-focus image generation. Blurry regions (yellow boxes) in the original image look clearer in the all-in-focus image.}
\label{fig:AllInFocus}
\vspace{-0.25in}

\begin{center}
\begin{tabular}{@{}c@{\hskip 0.001\linewidth}c@{\hskip 0.001\linewidth}c@{\hskip 0.001\linewidth}c@{\hskip 0.001\linewidth}c}
\includegraphics[width=0.199\linewidth]{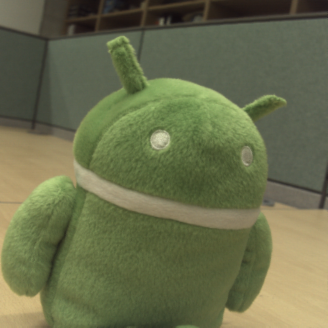} &
\includegraphics[width=0.199\linewidth]{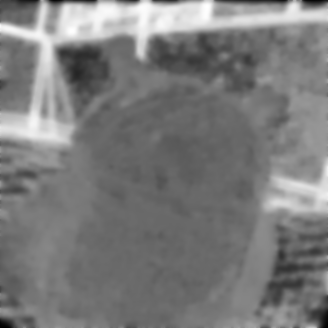} &
\includegraphics[width=0.199\linewidth]{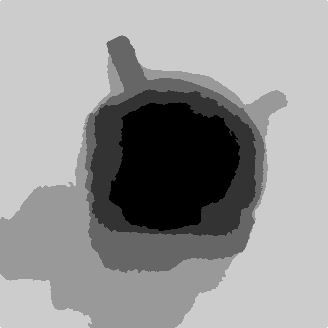} &
\includegraphics[width=0.199\linewidth]{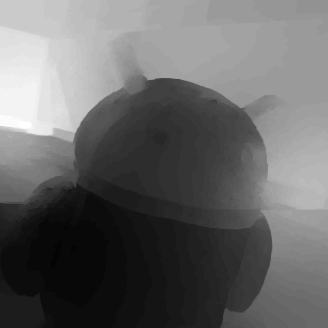} &
\includegraphics[width=0.199\linewidth]{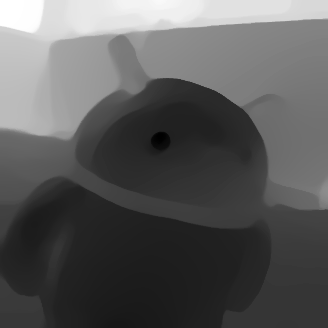} \\
(a) & (b) & (c) & (d) & (e)
\end{tabular}
\end{center}
\vspace{-0.25in}
\caption{Depth estimation. (a) Input image. (b) Tao \emph{et al.}~\cite{tao2013depth}. (c) Jeon \emph{et al.}~\cite{jeon2015accurate}. (d) Liu \emph{et al.}~\cite{LiuSL15} (e) Ours. Our depth map looks reasonable compared to those in the other works.}
\label{fig:DepthEstimation}
\vspace{-0.25in}
\end{figure}

\subsection{Blur Detection Dataset}
We verify the reliability and robustness of our algorithm using a blur detection dataset~\cite{shi2014discriminative}. The dataset used contains 704 natural images containing blurry and sharp regions and corresponding binary blurry region masks manually segmented by a human. We extract $15\times15$ patches on strong edges and $27\times27$ patches on weak edges. We set $n_{D} = n_{G} = n_{S} = 25$ for large patches, $n_{D} = n_{G} = n_{S} = 13$ for small patches, $\sigma_{s} = \sigma_{r} = 100.0$, $\sigma_{c} = 1.0$, $\epsilon = 1e^{-5}$ and $\gamma = 0.005$ for all experiments. We compare our algorithm to the results of Shi \emph{et al.}~\cite{shi2014discriminative}, Shi \emph{et al.}~\cite{shi2015just} and Zhuo and Sim~\cite{zhuo2011defocus}. Because the blur detection dataset contains only binary masks, quantitative results are obtained using binary blurry region masks from each algorithm. For binary segmentation, we apply a simple thresholding method to the full defocus map. The threshold value $\tau$ is determined as follows:
\begin{equation}
\tau = \alpha v_{max} + (1-\alpha)v_{min},
\label{eq:Threshold}
\end{equation}
where $v_{max}$ and $v_{min}$ denote the maximum and the minimum values of the full defocus map, and $\alpha$ is a user parameter. We set $\alpha = 0.3$ for the experiments empirically and this value works reasonably well. \figref{fig:QuantitativeComparison} shows the segmentation accuracies and the precision-recall curves, and \figref{fig:Results} shows the results of the different algorithms. The segmentation accuracies are obtained from the ratio of the number of pixels correctly classified to the total number of pixels. Precision-Recall curves can be calculated by adjusting $\tau$ from $\sigma_{1}$ to $\sigma_{L}$.

Our algorithm shows better results than the state-of-the-art methods quantitatively and qualitatively. In the homogeneous regions of an image, sufficient textures for defocus estimation do not exist. The results of ~\cite{shi2014discriminative} and ~\cite{shi2015just}, therefore, show some erroneous classification results in such regions. Our algorithm and ~\cite{zhuo2011defocus} can avoid this problem with the help of sparse patch extraction on strong edges. In contrast, sparse map propagation can also lead to uncertain results in textured regions because the prior used for the propagation is based on the color line model~\cite{levin2008closed}, as shown in \figref{fig:Results} (e). An edge-preserving smoothed color image is adopted as a propagation prior in our algorithm, and it gives better results, as shown in Figures \ref{fig:Results} (f) and (g). In addition, we compare our algorithm with that of Shi \emph{et al.}~\cite{shi2015break}. We conducted a \textit{RelOrder}~\cite{shi2015break} evaluation and obtained a result of 0.1572 on their dataset, which is much better than their result of 0.1393\footnote{Different from the reported value in \cite{shi2015break} because we utilize our own implementation. The evaluation codes of \cite{shi2015break} have not been released.} of \cite{shi2015break}. Moreover, the segmentation accuracy of \cite{shi2015break} (53.30\%) is much lower than the accuracy of our algorithm because their method easily fails with a large amount of blur (\figref{fig:Results} (b)).

We also examined defocus maps from single and concatenated features qualitatively. As shown in Figures \ref{fig:ResultWithEachFeature} (b), (c) and (d), single hand-crafted features give unsatisfactory results compared to deep or concatenated features. Surprisingly, a deep feature alone (\figref{fig:ResultWithEachFeature} (e)) works quite well but gives a slightly moderate result compared to the concatenated features (See the solid blue boxes). The hand-crafted feature alone (\figref{fig:ResultWithEachFeature} (f)) also works nicely but there are several misclassifications (See the dashed red boxes). These features show complementary roles when they are concatenated. Certain misclassifications due to the hand-crafted feature are well handled by the deep feature, and the discriminative power of the deep feature was strengthened with the aid of the hand-crafted features, as shown in \figref{fig:ResultWithEachFeature} (g).

\subsection{Applications}
The estimated defocus maps can be used for various applications. We apply our defocus maps to the following applications and the results are quite pleasing.

\textbf{Defocus Blur Magnification} \ Our algorithm can be used for defocus blur magnification tasks. We can highlight the foreground by amplifying the blurriness of the background. Figure \ref{fig:BlurMagnification} shows an example of the defocus blur magnification. The foreground objects appear more prominent in the blur magnified image.

\textbf{All-in-focus Image Generation} \ Contrary to defocus blur magnification, we can deblur blurry regions in an image to obtain an all-in-focus image. Using the $\sigma$ values of each pixel in the defocus map, we generate corresponding Gaussian blur kernels and use them to deblur the image. We use the hyper-laplacian prior~\cite{krishnan2009fast} for non-blind deconvolution. Figure \ref{fig:AllInFocus} shows an example of the all-in-focus image generation. In accordance with the defocus map, we deblur the original image in a pixel-by-pixel manner. The blurry regions in the original image are restored considerably in the all-in-focus image (\figref{fig:AllInFocus} (b)).

\textbf{3-D Estimation from a Single Image} \ The amount of defocus is closely related to the depth of the corresponding point because the scaled defocus values can be regarded as pseudo-depth values if all of the objects are located on the same side of the focal plane. Because we need both defocused images and depth maps, we utilize light-field images, as there are numerous depth estimation algorithms and because digital refocusing can easily be done.  We decode light-field images using \cite{cho2013modeling}, and then generate a refocused image using \cite{tao2013depth}. Our algorithm is compared to algorithms for light-field depth estimation~\cite{jeon2015accurate, tao2013depth} and an algorithm for single image depth estimation~\cite{LiuSL15}. \figref{fig:DepthEstimation} shows an input image and depth map from each algorithm. Our depth map from a single image appears reasonable compared to those in the other works, which utilize correspondences between multiple images.

\begin{figure}
\begin{center}
\begin{tabular}{@{}c@{\hskip 0.001\linewidth}c@{\hskip 0.001\linewidth}c@{\hskip 0.001\linewidth}c@{\hskip 0.001\linewidth}c}
\includegraphics[width=0.199\linewidth]{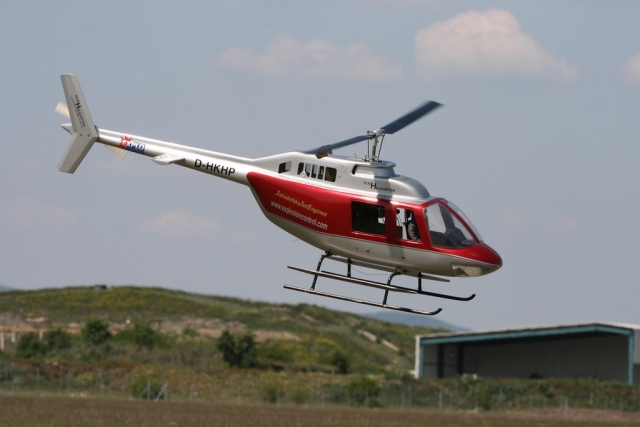} &
\includegraphics[width=0.199\linewidth]{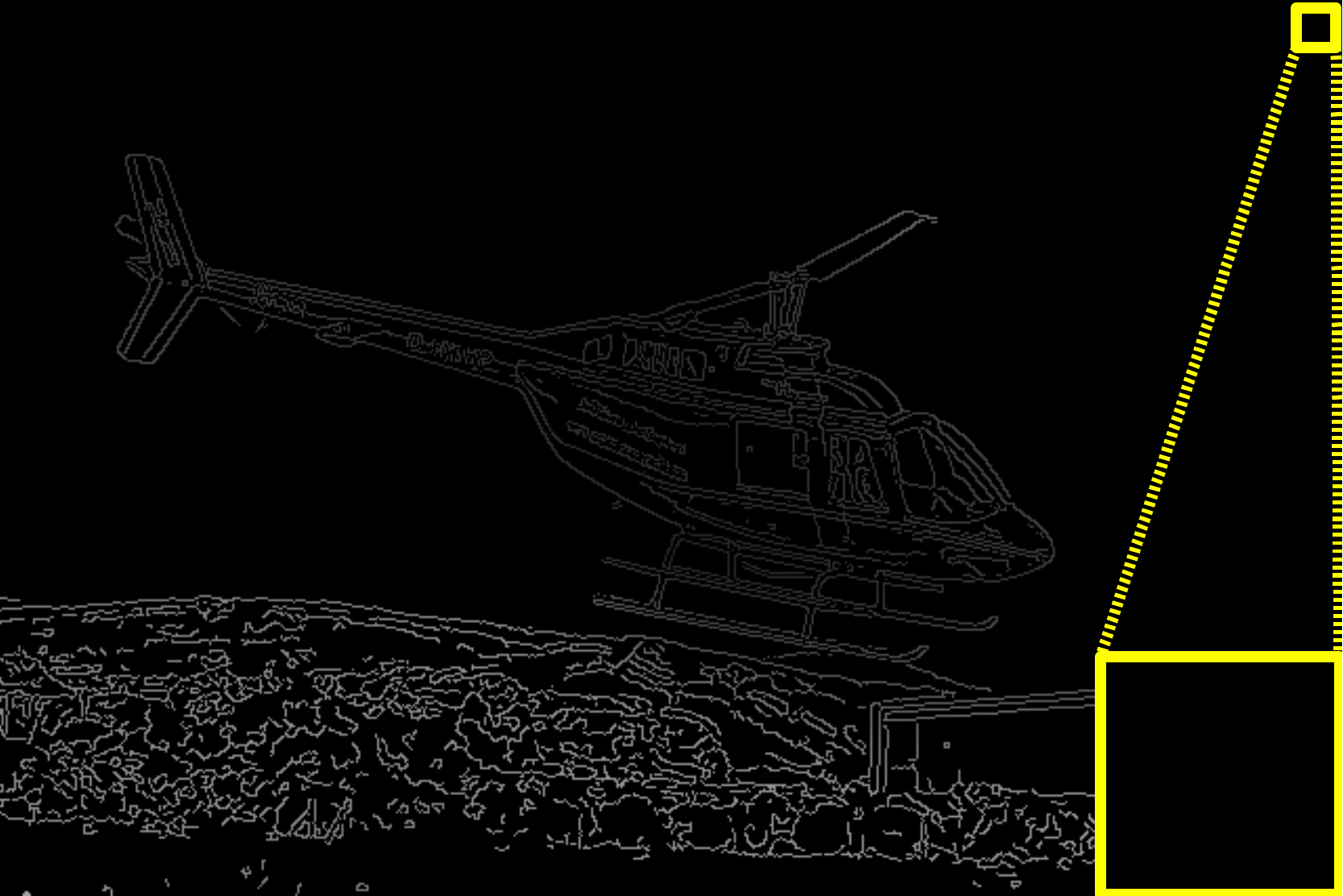} &
\includegraphics[width=0.199\linewidth]{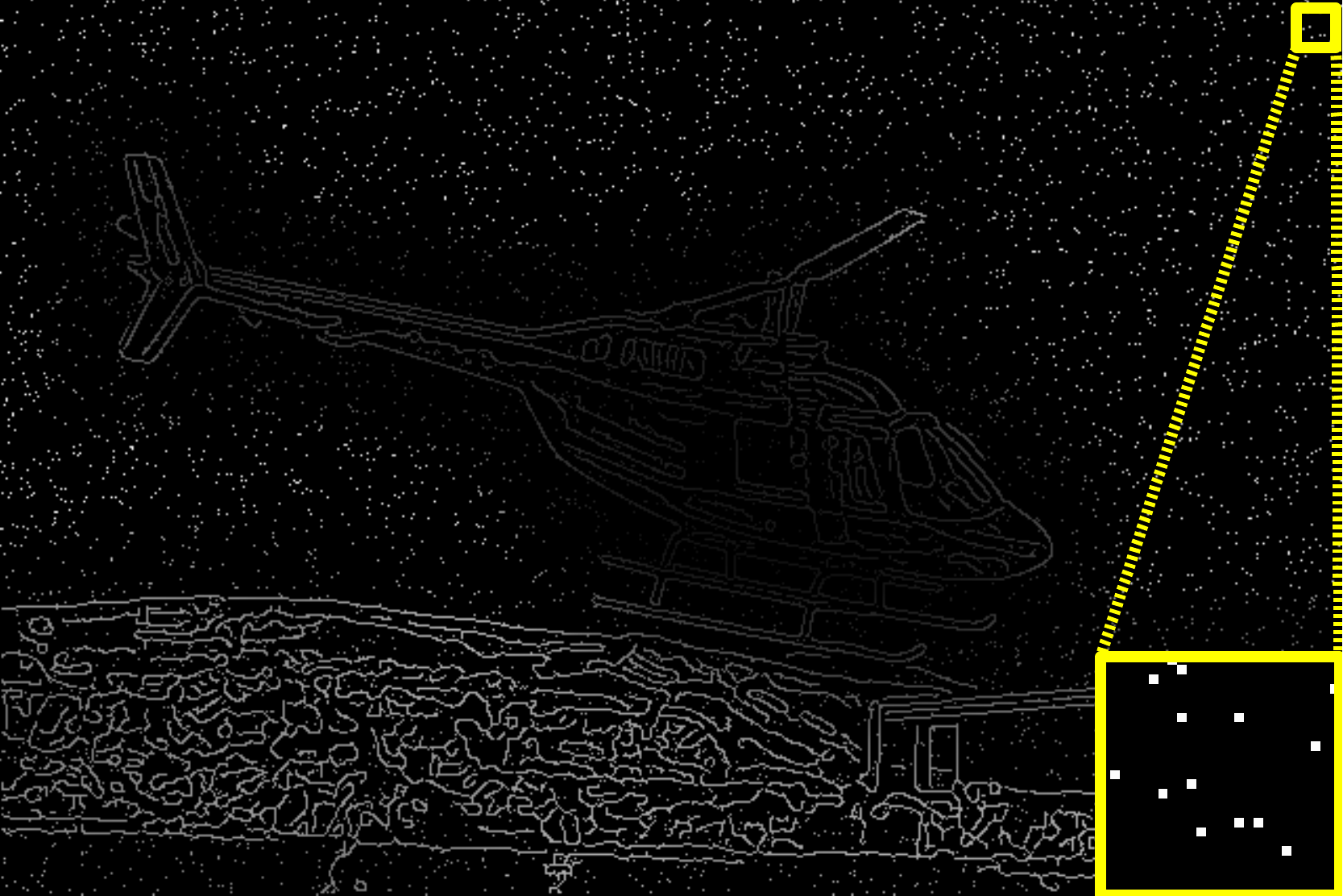} &
\includegraphics[width=0.199\linewidth]{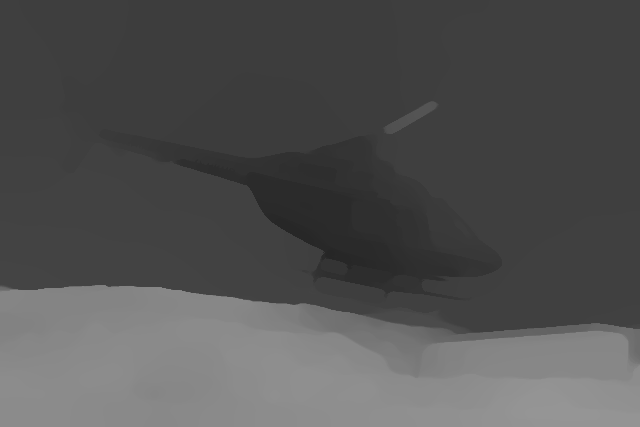} &
\includegraphics[width=0.199\linewidth]{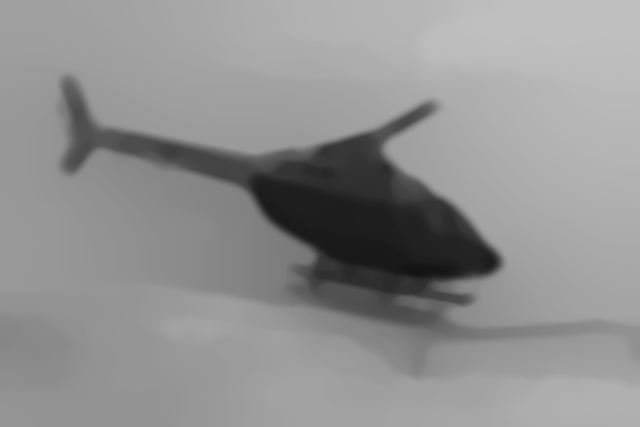} \\
{\small (a)} & {\small (b)} & {\small (c)} & {\small (d)} & {\small (e)}
\end{tabular}
\end{center}
\vspace{-0.25in}
\caption{Effects of additional random seed points. (a) Input image. (b) $I_{S}$. (c) $I_{S}$ + Seeds. (d) $I_{F}$ from (b). (e) $I_{F}$ from (c). Additional random seed points can effectively enhance the defocus accuracy in large homogeneous regions.}
\label{fig:RandomSeeds}
\vspace{-0.2in}
\end{figure}

\section{Conclusion}
\label{sec:Conclusion}
We have introduced a unified approach to combine hand-crafted and deep features and demonstrated their complementary effects for defocus estimation. A neural network classifier is shown to be able to capture highly non-linear relationships between each feature, resulting in high discriminative power. In order to reduce the patch scale dependency, multi-scale patches are extracted depending on the strength of the edges. The homogeneous regions in an image are well handled by a sparse defocus map, and the propagation process is guided by an edge-preserving smoothed image.  The performance of our algorithm is compared to those of the state-of-the-art algorithms. In addition, the potential for use in various applications is demonstrated.

One limitation of our algorithm is that we occasionally obtain incorrect defocus values in a large homogeneous area. This is due to the fact that there is no strong edge within such regions for defocus estimation. A simple remedy to address this problem involves the random addition of classification seed points in large homogeneous regions. \figref{fig:RandomSeeds} shows the effect of additional seed points. Additional random seed points effectively guide the sparse defocus map, causing it to be propagated correctly into large homogeneous areas.

For future works, we expect to develop fully convolutional network architectures for our task.

\section*{Acknowledgements}
This work was supported by the Technology Innovation Program (No. 2017-10069072) funded by the Ministry of Trade, Industry \& Energy (MOTIE, Korea).

{\small
\bibliographystyle{ieee}
\bibliography{dhde_arxiv}
}

\end{document}